\documentclass[10pt,journal,compsoc]{IEEEtran}

\usepackage{times}
\usepackage{epsfig}
\usepackage{graphicx}
\usepackage{amsmath}
\usepackage{amssymb}
\usepackage{dsfont}
\usepackage{siunitx}
\usepackage{subfigure}
\usepackage{multirow}
\usepackage{caption}
\usepackage{xcolor}
\usepackage{comment}

\usepackage[pagebackref=true,breaklinks=true,colorlinks,bookmarks=false]{hyperref}

\def\argmax{\mathop{\rm argmax}	}

\def\1{\mathds{1}}

\newcommand{\Y}{\mathcal{Y}}

\newcommand{\myparagraph}[1]{\vspace{6pt}\noindent{\bf #1}}
\newcommand{\modif}[1]{\textcolor{blue}{#1}}

\graphicspath{{./figures/}}

\begin{document}
\title{Zero-Shot Learning - A Comprehensive Evaluation of the Good, the Bad and the Ugly}

\author{Yongqin~Xian, Student Member, IEEE, Christoph~H.~Lampert,\\ Bernt~Schiele, Fellow, IEEE, and Zeynep~Akata, Member, IEEE
\IEEEcompsocitemizethanks{
\IEEEcompsocthanksitem Yongqin Xian and Bernt Schiele are with Max Planck
Institute for Informatics, Germany.
\IEEEcompsocthanksitem Christoph H. Lampert is with IST Austria (Institute of Science and Technology, Austria).
\IEEEcompsocthanksitem Zeynep Akata is with University of Amsterdam, Netherlands and with Max Planck
Institute for Informatics, Germany. 
}
\thanks{}}

\IEEEtitleabstractindextext{%

\begin{abstract}
Due to the importance of zero-shot learning, i.e. classifying images where there is a lack of labeled training data, the number of proposed approaches has recently increased steadily. We argue that it is time to take a step back and to analyze the status quo of the area. The purpose of this paper is three-fold. First, given the fact that there is no agreed upon zero-shot learning benchmark, we first define a new benchmark by unifying both the evaluation protocols and data splits of publicly available datasets used for this task. This is an important contribution as published results are often not comparable and sometimes even flawed due to, e.g. pre-training on zero-shot test classes. Moreover, we propose a new zero-shot learning dataset, the Animals with Attributes 2 (AWA2) dataset which we make publicly available both in terms of image features and the images themselves. Second, we compare and analyze a significant number of the state-of-the-art methods in depth, both in the classic zero-shot setting but also in the more realistic generalized zero-shot setting. Finally, we discuss in detail the limitations of the current status of the area which can be taken as a basis for advancing it. 
\end{abstract}

}

\maketitle
\IEEEdisplaynontitleabstractindextext

\section{Introduction}
\label{sec:intro}

\begin{figure}[t]
	\centering
        \includegraphics[width=\linewidth]{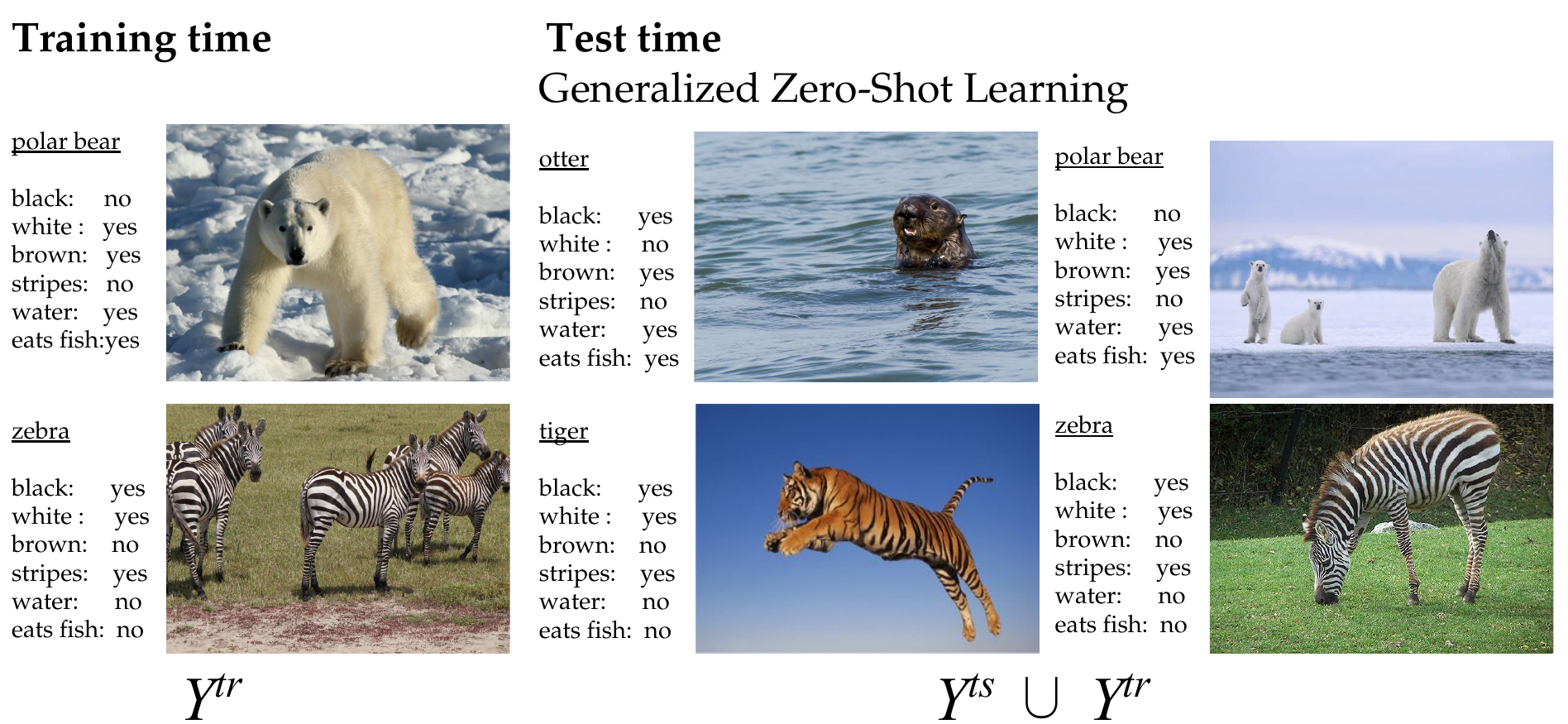}
	\caption{Zero-shot learning (ZSL) vs generalized zero-shot learning (GZSL): At training time, for both cases the images and attributes of the seen classes ($\Y^{tr}$) are available. At test time, in the ZSL setting, the learned model is evaluated only on unseen classes ($\Y^{ts}$) whereas in GZSL setting, the search space contains both training and test classes ($\Y^{tr}\cup \Y^{ts}$). To facilitate classification without labels, both tasks use some form of side information, e.g. attributes. The attributes are annotated per class, therefore the labeling cost is significantly reduced.}
	\label{fig:teaser}
\end{figure}

Zero-shot learning aims to recognize objects whose instances may not have been seen during training~\cite{LNH13,LEB08,RSS11,YA10,Xu17,Ding17}. The number of new zero-shot learning methods proposed every year has been increasing rapidly, i.e. the good aspects as our title suggests. Although each new method has been shown to make progress over the previous one, it is difficult to quantify this progress without an established evaluation protocol, i.e. the bad aspects. In fact, the quest for improving numbers has lead to even flawed evaluation protocols, i.e. the ugly aspects. Therefore, in this work, we propose to extensively evaluate a significant number of recent zero-shot learning methods in depth on several small to large-scale datasets using the same evaluation protocol both in zero-shot, i.e. training and test classes are disjoint, and the more realistic generalized zero-shot learning settings, i.e. training classes are present at test time. \autoref{fig:teaser} presents an illustration of zero-shot and generalized zero-shot learning tasks.

We benchmark and systematically evaluate zero-shot learning w.r.t. three aspects, i.e. methods, datasets and evaluation protocol. The crux of the matter for all zero-shot learning methods is to associate observed and non observed classes through some form of auxiliary information which encodes visually distinguishing properties of objects. Different flavors of zero-shot learning methods that we evaluate in this work are linear~\cite{FCSBDRM13,APHS13,ARWLS15,RT15} and nonlinear~\cite{XASNHS16,SGMN13} compatibility learning frameworks which have dominated the zero-shot learning literature in the past few years whereas an orthogonal direction is learning independent attribute~\cite{LNH13} classifiers and finally others~\cite{ZV15,CCGS16,NMBSSFCD14} propose a hybrid model between independent classifier learning and compatibility learning frameworks which have demonstrated improved results over the compatibility learning frameworks both for zero-shot and generalized zero-shot learning settings. 

We thoroughly evaluate the second aspect of zero-shot learning, by using multiple splits of several small, medium and large-scale datasets~\cite{PH12, CaltechUCSDBirdsDataset, LNH13, FEHF09, ImageNet}. Among these, the Animals with Attributes~(AWA1) dataset~\cite{LNH13} introduced as a zero-shot learning dataset with per-class attribute annotations, has been one of the most widely used datasets for zero-shot learning. However, as AWA1 images does not have the public copyright license, only some image features, i.e. SIFT~\cite{lowe2004distinctive}, DECAF~\cite{donahue2014decaf}, VGG19~\cite{simonyan2014very} of AWA1 dataset is publicly available, rather than the raw images. On the other hand, improving image features is a significant part of the progress both for supervised learning and for zero-shot learning. In fact, with the fast pace of deep learning, everyday new deep neural network models improve the ImageNet classification performance are being proposed.  Without access to images, those new DNN models can not be evaluated on AWA1 dataset. Therefore, with this work, we introduce the Animals with Attributes 2~(AWA2) dataset that has roughly the same number of images all with public licenses, exactly the same number of classes and attributes as the AWA1 dataset. We will make both ResNet~\cite{HZRS15} features of AWA2 images and the images themselves publicly available.  

We propose a unified evaluation protocol to address the third aspect of zero-shot learning which is one of the most important ones. We emphasize the necessity of tuning hyperparameters of the methods on a validation class split that is disjoint from training classes as improving zero-shot learning performance via tuning parameters on test classes violates the zero-shot assumption. We argue that per-class averaged top-1 accuracy is an important evaluation metric when the dataset is not well balanced with respect to the number of images per class. We point out that extracting image features via a pre-trained deep neural network (DNN) on a large dataset that contains zero-shot test classes also violates the zero-shot learning idea as image feature extraction is a part of the training procedure. Moreover, we argue that demonstrating zero-shot performance on small-scale and coarse grained datasets, i.e. aPY~\cite{FEHF09} is not conclusive. On the other hand, with this work we emphasize that it is hard to obtain labeled training data for fine-grained classes of rare objects recognizing which requires expert opinion. Therefore, we argue that zero-shot learning methods should be also evaluated on least populated or rare classes. We recommend to abstract away from the restricted nature of zero-shot evaluation and make the task more practical by including training classes in the search space, i.e. generalized zero-shot learning setting. Therefore, we argue that our work plays an important role in advancing the zero-shot learning field by analyzing the good and bad aspects of the zero-shot learning task as well as proposing ways to eliminate the ugly ones.

\section{Related Work}
\label{sec:related}

Early works of zero-shot learning~\cite{LNH13, HTS16, NMBSSFCD14, jayaraman2014zero, KKTH12} make use of the attributes within a two-stage approach to infer the label of an image that belong to one of the unseen classes. In the most general sense, the attributes of an input image are predicted in the first stage, then its class label is inferred by searching the class which attains the most similar set of attributes. For instance, DAP~\cite{LNH13} first estimates the posterior of each attribute for an image by learning probabilistic attribute classifiers. It then calculates the class posteriors and predicts the class label using MAP estimate. Similarly, \cite{HTS16} first learns a probabilistic classifier for each attribute. It then estimates the class posteriors through random forest which is able to handle unreliable attributes. IAP~\cite{LNH13} first predicts the class posterior of seen classes, then the probability of each class is used to calculate the attribute posteriors of an image. The class posterior of seen classes is predicted by a multi-class classifier. In addition, this two-stage approach have been extended to the case when attributes are not available. For example, following IAP~\cite{LNH13}, CONSE~\cite{NMBSSFCD14} first predicts seen class posteriors, then it projects image feature into the Word2vec~\cite{MSCCD13} space by taking the convex combination of top $T$ most possible seen classes. The two-stage models suffer from domain shift~\cite{FHXG15} between the intermediate task and target task, e.g. although the target task is to predict the class label, the intermediate task of  DAP is to learn attribute classifiers.

Recent advances in zero-shot learning directly learns a mapping from an image feature space to a semantic space. Among those, SOC~\cite{palatucci2009zero} maps the image features into the semantic space and then searches the nearest class embedding vector. ALE~\cite{APHS15} learns a bilinear compatibility function between the image and the attribute space using ranking loss. DeViSE~\cite{FCSBDRM13} also learns a linear mapping between image and semantic space using an efficient ranking loss formulation, and it is evaluated on the large-scale ImageNet dataset. SJE~\cite{ARWLS15} optimizes the structural SVM loss to learn the bilinear compatibility. On the other hand, ESZSL~\cite{RT15} uses the square loss to learn the bilinear compatibility and explicitly regularizes the objective w.r.t Frobenius norm. The $l_{2,1}$-based objective function of~\cite{QLSH16} suppresses the noise in the semantic space. \cite{BHJ16} embeds visual features into the attribute space, and then learns a metric to improve the consistency of the semantic embedding. Recently, SAE~\cite{kodirov2017semantic} proposed a semantic auto encoder to regularize the model by enforcing the image feature projected to the semantic space to be reconstructed.

Other zero-shot learning approaches learn non-linear multi-modal embeddings. LatEm~\cite{XASNHS16} extends the bilinear compatibility model of SJE~\cite{ARWLS15} to be a piecewise linear one by learning multiple linear mappings with the selection of which being a latent variable. CMT~\cite{SGMN13} uses a neural network with two hidden layers to learn a non-linear projection from image feature space to word2vec~\cite{MSCCD13} space. Unlike other works which build their embedding on top of fixed image features, \cite{lei2015predicting} trains a deep convolutional neural networks while learning a visual semantic embedding. Similarly, \cite{zhang2016learning} argues that the visual feature space is more discriminative than the semantic space, thus it proposes an end-to-end deep embedding model which maps semantic features into the visual space. \cite{changpinyo2016predicting} proposes a simple model by projecting class semantic representations into the visual feature space and performing nearest neighbor classifiers among those projected representations. The projection is learned through support vector regressor with visual exemplars of seen classes, i.e. class centroid in the feature space.

Embedding both the image and semantic features into another common intermediate space is another direction that zero-shot learning approaches adapt. SSE\cite{ZV15} uses the mixture of seen class proportions as the common space and argues that images belong to the same class should have similar mixture pattern. JLSE\cite{ZV16} maps visual features and semantic features into two separate latent spaces, and measures their similarity by learning another bilinear compatibility function. Furthermore, hybrid models~\cite{FXKG15,AMFS16,CCGS16, long2017zero} such as \cite{AMFS16} jointly embeds multiple text representations and multiple visual parts to ground attributes on different image regions. SYNC~\cite{CCGS16} constructs the classifiers of unseen classes by taking the linear combinations of base classifiers, which are trained in a discriminative learning framework.

While most of zero-shot learning methods learn the cross-modal mapping between the image and class embedding space with discriminative losses, there are a few generative models~\cite{VR17,LW17,MH16} that represent each class as a probability distribution. GFZSL~\cite{VR17} models each class-conditional distribution as a Gaussian and learns a regression function that maps a class embedding into the latent space. GLaP~\cite{LW17} assumes that each class-conditional distribution follows a Gaussian and generates virtual instances of unseen classes from the learned distribution. \cite{MH16} learns a multimodal mapping where class and image embeddings of categories are both represented by Gaussian distributions. 

Apart from the inductive zero-shot learning set-up where the model has no access to neither visual nor side-information of unseen classes, transductive zero-shot learning approaches~\cite{MES13,KXFG15, FHXG15,LGS15, li2015max, FS16, ZV16b} use visual or semantic information of both seen and unseen classes without having access to the label information. \cite{MES13} combines DAP and graph-based label propagation. \cite{KXFG15} uses the idea of domain adaptation frameworks. \cite{FHXG15} proposes hypergraph label propagation which allows to use multiple class embeddings. \cite{LGS15, li2015max, FS16} use semi-supervised learning based on max-margin framework.

In zero-shot learning, some form of side information is required to share information between classes so that the knowledge learned from seen classes is transfered to unseen classes. One popular form of side information is attributes, i.e. shared and nameable visual properties of objects. However, attributes usually require costly manual annotation. Thus, there has been a large group of studies~\cite{MGS14, rohrbach2010helps, ARWLS15, XASNHS16, FCSBDRM13, RALS16, QLSH16, lei2015predicting, ESE13, antol2014zero} which exploit other auxiliary information that reduces this annotation effort. \cite{mensink2012metric} does not use side information however it requires one-shot image of the novel class to perform nearest neighbor search with the learned metric. SJE~\cite{ARWLS15} evaluates four different class embeddings including attributes, word2vec~\cite{MSCCD13}, glove~\cite{PSM14} and wordnet hierarchy~\cite{WordNet}. On ImageNet, \cite{RSS11} leverages the wordnet hierarchy. \cite{RALS16} leverages the rich information of detailed visual descriptions obtained from novice users and improves the performance of attributes obtained from experts. Recently, \cite{KASB17} took a different approach and learned class embeddings using human gaze tracks showing that human gaze is class-specific.

Zero-shot learning has been criticized for being a restrictive set up as it comes with a strong assumption of the image used at prediction time can only come from unseen classes. Therefore, generalized zero-shot learning setting~\cite{SRSB13} has been proposed to generalize the zero-shot learning task to the case where both seen and unseen classes are used at test time. \cite{JSB14} argues that although ImageNet classification challenge performance has reached beyond human performance, we do not observe similar behavior of the methods that compete at the detection challenge which involves rejecting unknown objects while detecting the position and label of a known object. \cite{FCSBDRM13} uses label embeddings to operate on the generalized zero-shot learning setting whereas \cite{ZSYXLC16} proposes to learn latent representations for images and classes through coupled linear regression of factorized joint embeddings. On the other hand, \cite{BB16} introduces a new model layer to the deep net which estimates the probability of an input being from an unknown class and \cite{SGMN13} proposes a novelty detection mechanism. 

Although zero-shot vs generalized zero-shot learning evaluation works exist~\cite{RSS11, CCGS16b} in the literature, our work stands out in multiple aspects. For instance, \cite{RSS11} operates on the ImageNet 1K by using 800 classes for training and 200 for test. One of the most comprehensive works, \cite{CCGS16b} provides a comparison between five methods evaluated on three datasets including ImageNet with three standard splits and proposes a metric to evaluate generalized zero-shot learning performance.
On the other hand, we evaluate ten zero-shot learning methods on five datasets with several splits both for zero-shot and generalized zero-shot learning settings, provide statistical significance and robustness tests, and present other valuable insights that emerge from our benchmark. In this sense, ours is the most extensive evaluation of zero-shot and generalized zero-shot learning tasks in the literature.

\section{Evaluated Methods}
\label{sec:method}
We start by formalizing the zero-shot learning task and then we describe the zero-shot learning methods that we evaluate in this work. Given a training set $\mathcal{S} = \{(x_n, y_n), n=1...N\}$, with $y_n \in \mathcal{Y}^{tr}$ belonging to training classes, the task is to learn $f: \mathcal{X} \rightarrow \mathcal{Y}$ by minimizing the regularized empirical risk:
\begin{equation}\label{eq:compatibility}
\frac{1}{N}\sum_{n=1}^N L(y_n, f(x_n; W)) + \Omega(W)
\end{equation}
where $L(.)$ is the loss function and $\Omega(.)$ is the regularization term.
Here, the mapping $f: \mathcal{X} \rightarrow \mathcal{Y}$ from input to output embeddings is defined as:
\begin{equation}
f(x;W) = \argmax_{y\in\mathcal{Y}} F(x,y;W) 
\end{equation}
\noindent At test time, in zero-shot learning setting, the aim is to assign a test image to an unseen class label, i.e. $\mathcal{Y}^{ts}\subset \mathcal{Y}$ and in generalized zero-shot learning setting, the test image can be assigned either to seen or unseen classes, i.e. $\mathcal{Y}^{tr+ts}\subset \mathcal{Y}$ with the highest compatibility score. 

\subsection{Learning Linear Compatibility}
\label{subsec:compatibilty}
Attribute Label Embedding (ALE)~\cite{APHS15}, Deep Visual Semantic Embedding (DEVISE)~\cite{FCSBDRM13} and Structured Joint Embedding (SJE)~\cite{ARWLS15} use bi-linear compatibility function to associate visual and auxiliary information:
\begin{equation}\label{eq:comp}
F(x,y;W) = \theta(x)^T W \phi(y)
\end{equation}
\noindent where $\theta(x)$ and $\phi(y)$, i.e. image and class embeddings, both of which are given. $F(.)$ is parameterized by the mapping $W$, that is to be learned. Given an image, compatibility learning frameworks predict the class which attains the maximum compatibility score with the image.

Among the methods that are detailed below, ALE~\cite{APHS15}, DEVISE~\cite{FCSBDRM13} and SJE~\cite{ARWLS15} do early stopping to implicitly regularize Stochastic Gradient Descent~(SGD) while ESZSL~\cite{RT15} and SAE~\cite{kodirov2017semantic} explicitly regularize the embedding model as detailed below. 
In the following, we provide a unified formulation of these five zero-shot learning methods.

\myparagraph{DEVISE~\cite{FCSBDRM13}} uses pairwise ranking objective that is inspired from unregularized ranking SVM~\cite{J02}:
\begin{equation}\label{eq:rank}
\sum_{y\in \mathcal{Y}^{tr}}[\Delta(y_n,y) + F(x_n,y;W) - F(x_n,y_n;W)]_+
\end{equation}
\noindent where $\Delta(y_n,y)$ is equal to 1 if $y_n=y$, otherwise 0. The objective function is convex and is optimized by Stochastic Gradient Descent.

\myparagraph{ALE~\cite{APHS15}} uses the weighted approximate ranking objective~\cite{UBG09} for zero-shot learning in the following way:
\begin{equation}
\sum_{y\in \mathcal{Y}^{tr}} \frac{l_{r_{\Delta(x_n,y_n)}}}{r_{\Delta(x_n,y_n)}} [\Delta(y_n,y) + F(x_n,y;W) - F(x_n,y_n;W)]_+
\end{equation}
\noindent where $l_k= \sum_{i=1}^{k}\alpha_i$ and $r_{\Delta(x_n,y_n)}$ is defined as: 
\begin{equation}
\sum_{y \in \mathcal{Y}^{tr}} \mathds{1}(F(x_n,y;W) + \Delta(y_n,y) \geq F(x_n,y_n;W))
\end{equation} 
\noindent Following the heuristic in \cite{WBU11}, \cite{APHS15} selects $\alpha_i = 1/i$ which puts a high emphasis on the top of the rank list. 

\myparagraph{SJE~\cite{ARWLS15}} gives the full weight to the top of the ranked list and is inspired from the structured SVM~\cite{TJH05}:
\begin{equation}
[\max_{y\in \mathcal{Y}^{tr}}(\Delta(y_n,y) + F(x_n,y;W)) - F(x_n,y_n;W)]_+
\end{equation}
\noindent The prediction can only be made after computing the score against all the classifiers, i.e. so as to find the maximum violating class, which makes SJE less efficient than DEVISE and ALE.

\myparagraph{ESZSL~\cite{RT15}} applies a square loss to the ranking formulation and adds the following implicit regularization term to the unregularized risk minimization formulation:
\begin{equation}
\gamma\|W\phi(y)\|^2 + \lambda\|\theta(x)^TW\|^2 + \beta\|W\|^2
\end{equation}
\noindent where $\gamma, \lambda, \beta$ are regularization parameters. The first two terms bound the Euclidean norm of projected attributes in the feature space and projected image feature in the attribute space respectively. The advantage of this approach is that the objective function is convex and has a closed form solution.

\myparagraph{SAE~\cite{kodirov2017semantic}} also learns the linear projection from image embedding space to class embedding space, but it further constrains that the projection must be able to reconstruct the original image embedding. Similar to the linear auto-encoder, SAE optimizes the following objective:
\begin{equation}
\min_{W} ||\theta(x)-W^T \phi(y)||^2 + \lambda ||W\theta(x)-\phi(y)||^2, 
\label{eq:sae}
\end{equation}
\noindent where $\lambda$ is a hyperparameter to be tuned. The optimization problem can be transformed such that Bartels-Stewart algorithm~\cite{bartels1972solution} is able to solve it efficiently.

\subsection{Learning Nonlinear Compatibility}
Latent Embeddings (LATEM)~\cite{XASNHS16} and Cross Modal Transfer (CMT)~\cite{SGMN13} encode an additional non-linearity component to linear compatibility learning framework.

\myparagraph{LATEM~\cite{XASNHS16}} constructs a piece-wise linear compatibility:
\begin{equation}
F(x,y;W_i) = \max_{1\leq i\leq K}\theta(x)^T W_i \phi(y)
\end{equation}
\noindent where every $W_i$ models a different visual characteristic of the data and the selection of which matrix to use to do the mapping is a latent variable and $K$ is a hyperparameter to be tuned. LATEM uses the ranking loss formulated in~\autoref{eq:rank} and Stochastic Gradient Descent as the optimizer.

\myparagraph{CMT~\cite{SGMN13}} first maps images into a semantic space of words, i.e. class names, where a neural network with $\tanh$ nonlinearity learns the mapping:
\begin{equation}
\sum_{y\in \mathcal{Y}^{tr}} \sum_{x\in\mathcal{X}_y} \|\phi(y) - W_1 \tanh(W_2. \theta(x)\|^2
\end{equation}
\noindent where $(W_1,W_2)$ are weights of the two layer neural network. This is followed by a novelty detection mechanism that assigns images to unseen or seen classes. The novelty is detected either via thresholds learned using the embedded images of the seen classes or the outlier probabilities are obtained in an unsupervised way. As zero-shot learning assumes that test images are only from unseen classes, in our experiments when we refer to CMT, that means we do not use the novelty detection component. On the other hand, we name the CMT with novelty detection as CMT* when we apply it to the generalized zero-shot learning setting. 

\subsection{Learning Intermediate Attribute Classifiers}
Although Direct Attribute Prediction (DAP)~\cite{LNH13} and Indirect Attribute Prediction (IAP)~\cite{LNH13} have been shown to perform poorly compared to compatibility learning frameworks~\cite{APHS15}, we include them to our evaluation for being historically the most widely used methods in the literature. 

\myparagraph{DAP~\cite{LNH13}} learns probabilistic attribute classifiers and makes a class prediction by combining scores of the learned attribute classifiers. A novel image is assigned to one of the unknown classes using:
\begin{equation}
f(x) = \argmax_{c} \prod_{m=1}^M \frac{p(a_m^c|x)}{p(a_m^c)}.
\label{eq:dap}
\end{equation}
\noindent  with $M$ being the total number of attributes, $a^c_m$ is the m-th attribute of class $c$, $p(a_m^c|x)$ is the attribute probability given image $x$ which is obtained from the attribute classifiers whereas $p(a_m^c)$ is the attribute prior estimated by the empirical mean of attributes over training classes. We train binary classifiers with logistic regression that gives probability scores of attributes with respect to training classes.

\myparagraph{IAP~\cite{LNH13}} indirectly estimates attributes probabilities of an image by first predicting the probabilities of each training class, then multiplying the class attribute matrix. Once the attributes probabilities are obtained by the following equation: 
\begin{equation}
p(a_m|x) =  \sum_{k=1}^K p(a_m|y_k) p(y_k|x), 
\label{eq:iap}
\end{equation}
\noindent where $K$ is the number of training classes,  $p(a_m|y_k)$ is the predefined class attribute and $p(y_k|x)$ is training class posterior from multi-class classifier, the Equation~\ref{eq:dap} is used to predict the class label for which we train a multi-class classifier on training classes with logistic regression.

\subsection{Hybrid Models}

Semantic Similarity Embedding (SSE)~\cite{ZV15}, Convex Combination of Semantic Embeddings (CONSE)~\cite{NMBSSFCD14} and Synthesized Classifiers (SYNC)~\cite{CCGS16} express images and semantic class embeddings as a mixture of seen class proportions, hence we group them as hybrid models.

\myparagraph{SSE~\cite{ZV15}} leverages similar class relationships both in image and semantic embedding space. An image is labeled with:
\begin{equation}
\argmax_{u\in \mathcal{U}} \pi(\theta(x))^T \psi(\phi(y_u))
\end{equation}
\noindent where $\pi, \psi$ are mappings of class and image embeddings into a common space defined by the mixture of seen classes proportions. Specifically, $\psi$ is learned by sparse coding and $\pi$ is by class dependent transformation.

\myparagraph{CONSE~\cite{NMBSSFCD14}} learns the probability of a training image belonging to a training class:
\begin{equation}
f(x,t) = \argmax_{y\in\mathcal{Y}^{tr}}p_{tr}(y|x)
\end{equation}
\noindent where $y$ denotes the most likely training label ($t$=1) for image $x$. Combination of semantic embeddings ($s$) is used to assign an unknown image to an unseen class:
\begin{equation}
\frac{1}{Z} \sum_{i=1}^T p_{tr}(f(x,t)|x) . s(f(x,t))
\end{equation}
\noindent where $Z = \sum_{i=1}^T p_{tr}(f(x,t)|x)$, $f(x,t)$ denotes the t$^{th}$ most likely label for image $x$ and $T$ controls the maximum number of semantic embedding vectors.

\myparagraph{SYNC~\cite{CCGS16}} learns a mapping between the semantic class embedding space and a model space. In the model space, training classes and a set of phantom classes form a weighted bipartite graph. The objective is to minimize distortion error:
\begin{equation}\label{eq:H}
\min_{w_c} \|w_c - \sum_{r=1}^R s_{cr} v_r \|_2^2.
\end{equation}
\noindent Semantic and model spaces are aligned by embedding classifiers of real classes ($w_c$) and classifiers  of phantom classes ($v_r$) in the weighted graph ($s_{cr}$). The classifiers for novel classes are constructed by linearly combining classifiers of phantom classes.

\myparagraph{GFZSL~\cite{VR17}} proposes a generative framework for zero-shot learning by modeling each class-conditional distribution as a multi-variate Gaussian with mean vector $\mu$ and diagonal covariance matrix $\sigma$. While the parameters of seen classes can be estimated by MLE, that of unseen classes are computed by learning the following two regression functions:
\begin{equation}\label{eq:regression1}
\mu_y = f_{\mu}(\phi(y)) \text{ and } \sigma_y = f_{\sigma}(\phi(y))
\end{equation}
%
%
with an image $x$, its class is predicted by searching the class with the maximum probability, i.e. $\argmax_y p(x|\sigma_y, \mu_y)$.

\subsection{Transductive Zero-Shot Learning Setting}
In zero-shot learning, transductive setting~\cite{chapelle2009semi,zhou2004learning} implies that unlabeled images from unseen classes are available during training. Using unlabeled images are expected to improve performance as they possibly contain useful latent information of unseen classes. Here, we mainly focus on two state-of-the-art transductive approaches\cite{VR17,YG17} and show how to extend ALE~\cite{APHS15} into the transductive learning setting.

\myparagraph{GFZSL-tran~\cite{VR17}} uses an Expectation-Maximization (EM) based procedure that alternates between inferring the labels of unlabeled examples of unseen classes and using the inferred labels to update the parameter estimates of unseen class distributions. Since the class-conditional distribution is assumed to be Gaussian, this procedure is equivalent to repeatedly estimating a Gaussian Mixture Model~(GMM) with the unlabeled data from unseen classes and use the inferred class labels to re-estimate the GMM.

\myparagraph{DSRL~\cite{YG17}} proposes to simultaneously learn image features with non-negative matrix factorization and align them with their corresponding class attributes. This step gives us an initial prediction score matrix $S_0$ in which each row is one instance and indicates the prediction scores for all unseen classes.
To improve the prediction score matrix by transductive learning, a graph-based label propagation algorithm is applied. Specifically, a KNN graph is constructed with the projected instances of unseen classes in the class embedding space,    
\begin{equation}
M_{ij}=
\left\{
    \begin{array}{ll}
            exp(-\frac{d(x_i, x_j)}{2\sigma^2})  & \mbox{if $i\in$ KNN($j$) or $j\in$ KNN($i$)}  \\
            0  & \mbox{otherwise}
    \end{array}
    \right.
\end{equation}
where KNN($i$) denotes the k-nearest neighbor of $i$-th instance and $d(x_i, x_j)$ measures the Euclidean distance between $x_i$ and $x_j$. Given the affinity matrix $M$, a normalized Laplacian matrix $L$ can be computed as $L=Q^{-1/2} M Q^{-1/2}$ where $Q$ is a diagonal matrix with $Q_{ii}=\sum_jM_{ij}$. Finally, the standard label propagation~\cite{fujiwara2014efficient} gives the closed-form solution:
\begin{equation}
S = (I - \alpha L)^{-1} \times S_0
\end{equation}
where $\alpha\in [0,1]$ is a regularization trade-off parameter and $S$ is the score matrix. 
The class label of an instance is predicted by searching the class with the highest score, i.e. 
%
$\argmax_y S_{iy}$.

\myparagraph{ALE-tran} Any compatibility learning method that explicitly learns cross-modal mapping from image feature space to class embedding space can be extended to transductive setting following the label propagation procedure of DSRL~\cite{YG17}. Taking the ALE~\cite{APHS15} as an example, after learning the linear mapping $W$, instances of unseen classes can be projected into the class embedding space and a score matrix $S_0$ can be computed similarly.    


\section{Datasets}
\label{sec:datasets}
Among the most widely used datasets for zero-shot learning, we select two coarse-grained, one small (aPY~\cite{FEHF09}) and one medium-scale (AWA1~\cite{LNH13}), and two fine-grained, both medium-scale, datasets (SUN~\cite{PH12}, CUB~\cite{CaltechUCSDBirdsDataset}) with attributes and one large-scale dataset (ImageNet~\cite{ImageNet}) without. Here, we consider  between $10K$ and $1M$ images, and, between $100$ and $1K$ classes as medium-scale. Details of dataset statistics in terms of the number of images, classes, attributes for the attribute datasets are in~\autoref{tab:datasets}.
Furthermore, we introduce our Animals With Attributes 2 (AWA2) dataset and position it with respect to existing datasets.

\subsection{Attribute Datasets} 
Attribute Pascal and Yahoo (aPY)~\cite{FEHF09} is a small-scale coarse-grained dataset with $64$ attributes. Among the total number of $32$ classes, $20$ Pascal classes are used for training (we randomly select $5$ for validation) and $12$ Yahoo classes are used for testing.
The original Animals with Attributes (AWA1)~\cite{LNH13} is a coarse-grained dataset that is medium-scale in terms of the number of images, i.e. $30,475$ and small-scale in terms of number of classes, i.e. $50$ classes. \cite{LNH13} introduces a standard zero-shot split with $40$ classes for training (we randomly select $13$ classes for validation) and $10$ classes for testing. AWA1 has $85$ attributes. 
Caltech-UCSD-Birds 200-2011 (CUB)~\cite{CaltechUCSDBirdsDataset} is a fine-grained and medium scale dataset with respect to both number of images and number of classes, i.e. $11,788$ images from $200$ different types of birds annotated with $312$ attributes. \cite{APHS15} introduces the first zero-shot split of CUB with $150$ training ($50$ validation classes) and $50$ test classes. 
SUN~\cite{PH12} is a fine-grained and medium-scale dataset with respect to both number of images and number of classes, i.e. SUN contains $14340$ images coming from $717$ types of scenes annotated with $102$ attributes. Following~\cite{LNH13} we use $645$ classes of SUN for training (we randomly select $65$ classes for validation) and $72$ classes for testing. 

\begin{figure}[t]
	\vspace{-3mm}
    \includegraphics[width=\linewidth, trim=0 0 0 0,clip]{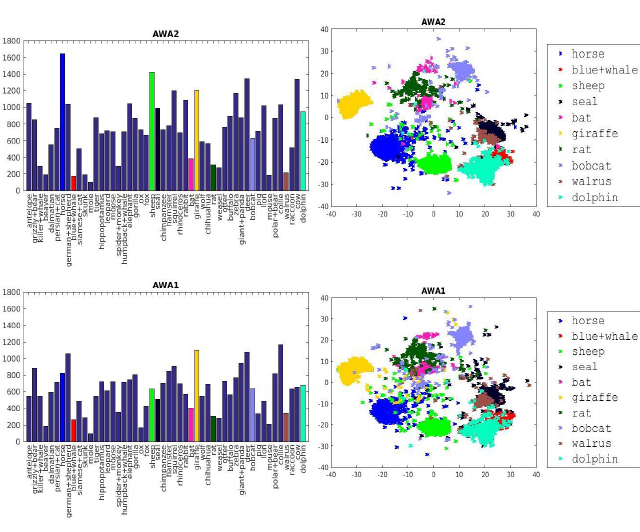} 
    \caption{Comparing AWA1~\cite{LNH13} and our AWA2 in terms of number of images (Left) and t-SNE embedding of the image features (the embedding is learned on AWA1 and AWA2 simultaneously, therefore the figures are comparable). AWA2 follows a similar distribution as AWA1 and it contains more examples.} \vspace{-3mm}
	\label{fig:tsne}
\end{figure}

\myparagraph{Animals with Attributes2 (AWA2) Dataset.}
One disadvantage of AWA1 dataset is that the images are not publicly available. As having highly descriptive image features is an important component for zero-shot learning, in order to enable vision research on the objects of the AWA1 dataset, we introduce the Animals with Attributes2~(AWA2) dataset. Following \cite{LNH13}, we collect $37,322$ images for the $50$ classes of AWA1 dataset from public web sources, i.e. Flickr, Wikipedia, etc., making sure that all images of AWA2 have free-use and redistribution licenses and they do not overlap with images of the original Animal with Attributes dataset. The AWA2 dataset uses the same 50 animal classes as AWA1 dataset, similarly the $85$ binary and continuous class attributes are common. In total, AWA2 has $37,322$ images compared to $30,475$ images of AWA1. On average, each class includes $746$ images where the least populated class, i.e. mole, has $100$ and the most populated class, i.e. horse has $1645$ examples. Some example images from \textit{polar bear, zebra, otter} and \textit{tiger} classes along with sample attributes  from our AWA2 dataset are shown in \autoref{fig:teaser}.

In \autoref{fig:tsne}, we provide some statistics on the AWA2 dataset in comparison with the AWA1 dataset in terms of the number of images and also the distribution of the image features. Compared to AWA1, our proposed AWA2 dataset contains more images, e.g. \textit{horse} and \textit{dolphin} among the test classes, \textit{antelope} and \textit{cow} among the training classes. Moreover, the t-SNE embedding of these test classes with more training data, e.g. \textit{horse}, \textit{dolphin}, \textit{seal} etc. shows that AWA2 leads to slightly more visible clusters of ResNet features. The images, their labels and ResNet features of our AWA2 are publicly available in \url{http://cvml.ist.ac.at/AwA2}. 

\subsection{Large-Scale ImageNet} 
We also evaluate the performance of methods on the large scale ImageNet~\cite{ImageNet} which contains a total of $14$ million images from $21$K classes, each one labeled with one label, and the classes are hierarchically related as ImageNet follows the WordNet~\cite{WordNet}. 

ImageNet is a natural fit for zero-shot and generalized zero-shot learning as there is a large class imbalance problem. Moreover, ImageNet is diverse in terms of granularity, i.e. it contains a collection of fine-grained datasets, e.g. different vehicle types, as well as coarse-grained datasets. The highest populated class contains $3,047$ images whereas there are many classes that contains only a single image. A balanced subset of ImageNet with $1$K classes containing about $1000$ images each is used to train CNNs. 

Previous works~\cite{RSS11} proposed to split the balanced subset of $1$K classes into $800$ training and $200$ test classes. In this work, from the total of $21$K classes, we use $1$K classes for training (among which we use $200$ classes for validation) and the test split is either all the remaining 20K classes or a subset of it, e.g. we determine these subsets based on the hierarchical distance between classes and the population of classes. The details of these splits are provided in the following section.

{
\renewcommand{\arraystretch}{1}
\begin{table*}[t]
  \begin{tabular}{ l c c c  c c c | c c c c c  c c c c|}
    & & & & \multicolumn{3}{c}{\multirow{3}{*}{\textbf{Number of Classes}}}& \multicolumn{9}{c}{\textbf{Number of Images}} \\
    \cline{8-16}
    & & & &  & & & & \multicolumn{4}{c}{\textbf{At Training Time}} & \multicolumn{4}{c|}{\textbf{At Evaluation Time}} \\
    & & & & & & & & \multicolumn{2}{c}{\textbf{SS}} & \multicolumn{2}{c}{\textbf{PS}} & \multicolumn{2}{c}{\textbf{SS}} & \multicolumn{2}{c|}{\textbf{PS}} \\
    \textbf{Dataset} & \textbf{Size} & \textbf{Granularity} & \textbf{Att} & $\mathcal{Y}$ & $\mathcal{Y}^{tr}$ & $\mathcal{Y}^{ts}$ & \textbf{Total} & $\mathcal{Y}^{tr}$ & $\mathcal{Y}^{ts}$ & $\mathcal{Y}^{tr}$ & $\mathcal{Y}^{ts}$ & $\mathcal{Y}^{tr}$ & $\mathcal{Y}^{ts}$ & $\mathcal{Y}^{tr}$ & $\mathcal{Y}^{ts}$ \\     \hline
    SUN~\cite{PH12} & medium & fine & 102 & 717 & $580$ + $65$ & 72 & 14340 & $12900$ & $0$ & $10320$ & $0$ &$0$ & $1440$ & $2580$ & $1440$ \\
    CUB~\cite{CaltechUCSDBirdsDataset} & medium & fine  & 312 & 200 & 100 + 50 & 50 & 11788 & $8855$ & $0$ & $7057$ & $0$ &$0$ & $2933$ & $1764$ & $2967$ \\
    AWA1~\cite{LNH13} & medium & coarse  & 85 & 50 & 27 + 13 & 10 & 30475 & $24295$ & $0$ & $19832$ & $0$ &$0$ & $6180$ & $4958$ & $5685$ \\
        AWA2 & medium & coarse  & 85 & 50 & 27 + 13 & 10 & 37322 & $30337$ & $0$ & $23527$ & $0$ &$0$ & $6985$ & $5882$ & $7913$ \\
    aPY~\cite{FEHF09} & small & coarse & 64 & 32 & $15$ + $5$ & 12 & 15339 & $12695$ & $0$ & $5932$ & $0$ &$0$ & $2644$ & $1483$ & $7924$ \\ 
    \hline %
  \end{tabular}
\caption{Statistics for SUN~\cite{PH12}, CUB~\cite{CaltechUCSDBirdsDataset}, AWA1~\cite{LNH13}, proposed AWA2, aPY~\cite{FEHF09} in terms of size, granularity, number of attributes, number of classes in $\mathcal{Y}^{tr}$ and $\mathcal{Y}^{ts}$, number of images at training and test time for standard split (SS) and our proposed splits (PS).}
\label{tab:datasets}
\end{table*}
}

\section{Evaluation Protocol}
In this section, we provide several components of previously used and our proposed ZSL and GZSL evaluation protocols, e.g. image and class encodings, dataset splits and the evaluation criteria\footnote{Our benchmark is in: \url{http://www.mpi-inf.mpg.de/zsl-benchmark}}. 

\subsection{Image and Class Embedding} 
We extract image features, namely image embeddings, from the entire image for SUN, CUB, AWA1, our AWA2 and ImageNet, with no image pre-processing. For aPY, following the original publication in~\cite{FEHF09}, we crop the images from bounding boxes. Our image embeddings are $2048$-dim top-layer pooling units of the $101$-layered ResNet~\cite{HZRS15} as we found that it performs better than $1,024$-dim top-layer pooling units of GoogleNet~\cite{SLJSRAEVR15}. We use the original ResNet-$101$ that is pre-trained on ImageNet with $1$K classes, i.e. the balanced subset, and we do not fine-tune it for any of the mentioned datasets. In addition to the ResNet features, we re-evaluate all methods with their published image features. 

In zero-shot learning, class embeddings are as important as image features. As class embeddings, for aPY, AWA1, AWA2, CUB and SUN, we use the per-class attributes between values $0$ and $1$ that are provided with the datasets as binary attributes have been shown~\cite{APHS15} to be weaker than continuous attributes. For ImageNet as attributes of $21$K classes are not available, we use Word2Vec~\cite{MSCCD13} trained on Wikipedia provided by~\cite{CCGS16}. Note that an evaluation of class embeddings is out of the scope of this paper. We refer the reader to~\cite{ARWLS15} for more details on the topic.

\subsection{Dataset Splits}
Zero-shot learning assumes disjoint training and test classes. Hence, as deep neural network (DNN) training for image feature extraction is actually a part of model training, the dataset used to train DNNs, e.g. ImageNet, should not include any of the test classes. However, we notice from the standard splits (SS) of aPY and AWA1 datasets that 7 aPY test classes out of 12 (monkey, wolf, zebra, mug, building, bag, carriage), 6 AWA1 test classes out of 10 (chimpanzee, giant panda, leopard, persian cat, pig, hippopotamus), are among the 1K classes of ImageNet, i.e. are used to pre-train ResNet. On the other hand, the mostly widely used splits, i.e. we term them as standard splits (SS), for SUN from~\cite{LNH13} and CUB from~\cite{APHS13} shows us that 1 CUB test class out of 50 (Indigo Bunting), and 6 SUN test classes out of 72 (restaurant, supermarket, planetarium, tent, market, bridge), are also among the 1K classes of ImageNet. 

We noticed that the accuracy for all methods on those overlapping test classes are higher than others. Therefore, we propose new dataset splits, i.e. proposed splits (PS), insuring that none of the test classes appear in ImageNet 1K, i.e. used to train the ResNet model. 
We present the differences between the standard splits (SS) and the proposed splits (PS) in~\autoref{tab:datasets}. While in SS and PS no image from test classes is present at training time, at test time our PS includes images from training classes. We designed the PS this way as evaluating accuracy on both training and test classes is crucial to show the generalization of the methods. 

For SUN, CUB, AWA1, aPY, and our proposed AWA2 dataset, for measuring the significance of the results, we propose 3 different splits of $580$, $100$, $27$, $15$ and $27$ training classes respectively while keeping $72$, $50$, $10$, $12$ and $10$ test classes the same. It is important to perform hyperparameter search on a disjoint set of validation set of $65$, $50$, $13$, $5$ and $13$ classes respectively. We keep the number of classes the same for SS and PS, however we choose different classes while making sure that the test classes do not overlap with the $1$K training classes of ImageNet. \modif{Note that we introduce Proposed Split Version 2.0\footnote{\url{http://www.mpi-inf.mpg.de/zsl-benchmark}}.}

ImageNet provides possibilities of constructing several zero-shot evaluation splits. Following~\cite{CCGS16}, our first two standard splits consider all the classes that are 2-hops and 3-hops away from the original 1K classes according to the ImageNet label hierarchy, corresponding to $1509$ and $7678$ classes. This split measures the generalization ability of the models with respect to the hierarchical and semantic similarity between classes. As discussed in the previous section, another characteristic of ImageNet is the imbalanced sample size. Therefore, our proposed split considers 500, 1K and 5K most populated classes among the remaining 21K classes of ImageNet with approximately $1756$, $1624$ and $1335$ images per class on average. Similarly, we consider 500, 1K and 5K least-populated classes in ImageNet which correspond to most fine-grained subsets of ImageNet with approximately $1$, $3$ and $51$ images per class on average. We measure the generalization of methods to the entire ImageNet data distribution by considering a final split of all the remaining approximately $20$K classes of ImageNet with at least $1$ image per-class, i.e. approximately $631$ images per class on average.

\subsection{Evaluation Criteria}
Single label image classification accuracy has been measured with Top-1 accuracy, i.e. the prediction is accurate when the predicted class is the correct one. If the accuracy is averaged for all images, high performance on densely populated classes is encouraged. However, we are interested in having high performance also on sparsely populated classes. Therefore, we average the correct predictions independently for each class before dividing their cumulative sum w.r.t the number of classes, i.e. we measure average per-class top-1 accuracy in the following way:
\begin{equation}
acc_{\Y} = \frac{1}{\|\Y\|} \sum_{c=1}^{\|\Y\|} \frac{\text{\# correct predictions in c}}{\text{\# samples in c}}
\end{equation}
In the generalized zero-shot learning setting, the search space at evaluation time is not restricted to only test classes ($\Y^{ts}$), but includes also the training classes ($\Y^{tr}$), hence this setting is more practical. As with our proposed split at test time we have access to some images from training classes, after having computed the average per-class top-1 accuracy on training and test classes, we compute the harmonic mean of training and test accuracies:
\begin{equation}
H = \frac{2*acc_{\Y^{tr}}*acc_{\Y^{ts}}}{acc_{\Y^{tr}}+acc_{\Y^{ts}}}
\end{equation}
\noindent where $acc_{\mathcal{Y}^{tr}}$ and $acc_{\mathcal{Y}^{ts}}$ represent the accuracy of images from seen ($\mathcal{Y}^{tr}$), and images from unseen ($\mathcal{Y}^{ts}$) classes respectively. We choose harmonic mean as our evaluation criteria and not arithmetic mean because in arithmetic mean if the seen class accuracy is much higher, it effects the overall results significantly. Instead, our aim is high accuracy on both seen and unseen classes.

{
\setlength{\tabcolsep}{4pt}
\renewcommand{\arraystretch}{1}
\begin{table}[t]
 \centering
      \begin{tabular}{l c c c c c c c c}
& \multicolumn{2}{c}{\textbf{SUN}} & \multicolumn{2}{c }{\textbf{CUB}} & \multicolumn{2}{c }{\textbf{AWA1}} & \multicolumn{2}{c }{\textbf{aPY}} \\
      \textbf{Model} & \textbf{R} & \textbf{O}   & \textbf{R} & \textbf{O} & \textbf{R} & \textbf{O} & \textbf{R} & \textbf{O}\\     
     \hline
     DAP~\cite{LNH13} & $22.1$ & $22.2$ & $-$ & $-$ & $41.4$ & $41.4$ & $19.1$ & $19.1$\\
     SSE~\cite{ZV15} & $83.0$ & $82.5$ & $44.2$ & $30.4$ & $64.9$ & $76.3$ & $45.7$ & $46.2$ \\
     LATEM~\cite{XASNHS16} & $-$ & $-$ & $45.1$ & $45.5$ & $71.2$ & $71.9$ & $-$ & $-$\\
     SJE~\cite{ARWLS15} & $-$ & $-$ & $50.1$ & $50.1$ & $67.2$ & $66.7$ & $-$ & $-$\\
     ESZSL~\cite{RT15} & $64.3$ & $65.8$ & $-$ & $-$ & $48.0$ & $49.3$ & $14.3$ & $15.1$\\
     SYNC~\cite{CCGS16} & $62.8$ & $62.8$ & $53.4$ & $53.4$ & $69.7$ & $69.7$ & $-$ & $-$\\ 
     SAE~\cite{kodirov2017semantic} & $-$ & $-$ & $-$ & $-$ & $84.7$ & $84.7$ & $-$ & $-$\\ 
     GFZSL~\cite{VR17} & $86.5$ & $86.5$ & $56.6$ & $56.5$ & $80.4$ & $80.8$ & $-$ & $-$\\    
     \hline
     GFZSL-tran~\cite{VR17} & $87.0$ & $87.0$ & $63.8$ & $63.7$ & $94.9$ & $94.3$ & $-$ & $-$\\ 
     DSRL~\cite{YG17} & $86.0$ & $85.4$ & $57.6$ & $57.1$ & $87.7$ & $87.2$ & $47.8$ & $51.3$\\ 
     \hline
   \end{tabular} 
\caption{Reproducing zero-shot results with methods that have a public implementation: O = Original results, R = Reproduced using provided image features and code. We measure top-1 accuracy in \%. $-$: image features are not provided in the original paper for this dataset. Top: ZSL, Bottom: transductive ZSL.}
\label{tab:reproduce}
\end{table}
}

{
\renewcommand{\arraystretch}{1}
\begin{table*}[t]
 \centering
   \begin{tabular}{l c c |c c |c c |c c |c c}
      & \multicolumn{2}{c }{\textbf{SUN}} & \multicolumn{2}{c }{\textbf{CUB}} & \multicolumn{2}{c }{\textbf{AWA1}} & \multicolumn{2}{c }{\textbf{AWA2}}  & \multicolumn{2}{c }{\textbf{aPY}} \\
      \textbf{Method} & \textbf{SS} & \textbf{PS} & \textbf{SS} & \textbf{PS} & \textbf{SS} & \textbf{PS} & \textbf{SS} & \textbf{PS} & \textbf{SS} & \textbf{PS} \\     
     \hline
     DAP~\cite{LNH13}  & $38.9$ & $39.9$ & $37.5$ & $40.0$ & $57.1$ & $44.1$ & $58.7$ & $46.1$ & $35.2$ & $33.8$ \\
     IAP~\cite{LNH13}  & $17.4$ & $19.4$ & $27.1$ & $24.0$ & $48.1$ & $35.9$ & $46.9$ & $35.9$ & $22.4$ & $36.6$ \\
     CONSE~\cite{NMBSSFCD14} & $44.2$ & $38.0$ & $36.7$ & $33.6$ & $63.6$ & $46.3$ & $67.9$ & $44.6$ & $25.9$ & $26.4$\\
     CMT~\cite{SGMN13} & $41.9$ & $40.1$ & $37.3$ & $34.6$ & $58.9$ & $39.5$ & $66.3$ & $37.9$ & $26.9$ & $28.0$\\
     SSE~\cite{ZV15} & $54.5$ & $51.5$ & $43.7$ & $43.9$ & $68.8$ & $60.1$ & $67.5$ & $61.0$ & $31.1$ & $35.0$\\
     LATEM~\cite{XASNHS16} & $56.9$ & $55.3$ & $49.4$ & $49.6$ & $74.8$ & $55.1$ & $68.7$ & $55.8$ & $34.5$ & $36.8$\\
     ALE~\cite{APHS15} & $59.1$ & $58.1$ & $53.2$ & $54.9$ & $78.6$ & $59.9$ & $80.3$ & $62.5$ & $30.9$ & $\mathbf{39.7}$ \\
     DEVISE~\cite{FCSBDRM13}  & $57.5$ & $56.5$ & $53.2$ & $52.0$ & $72.9$& $54.2$ & $68.6$ & $59.7$ & $35.4$ & $37.0$ \\
     SJE~\cite{ARWLS15} & $57.1$ & $52.7$ & $\mathbf{55.3}$ & $53.9$ & $76.7$ & $65.6$ & $69.5$ & $61.9$ & $32.0$ & $31.7$\\
     ESZSL~\cite{RT15}  & $57.3$ & $54.5$ & $55.1$ & $51.9$ & $74.7$ & $58.2$ & $75.6$ & $58.6$ & $34.4$ & $38.3$\\
     SYNC~\cite{CCGS16} & $59.1$ & $56.2$ & $54.1$ & $\mathbf{56.0}$ & $72.2$ & $51.8$ & $71.2$ & $49.3$ & $39.7$ & $23.9$ \\ 
     SAE~\cite{kodirov2017semantic} & $42.4$ & $40.3$ & $33.4$ & $33.3$ & $\mathbf{80.6}$ & $53.0$ & $\mathbf{80.7}$ & $54.1$ & $8.3$ & $8.3$ \\ 
     GFZSL~\cite{VR17} & $\mathbf{62.9}$ & $\mathbf{60.6}$ & $53.0$ & $49.3$ & $80.5$ & $\mathbf{68.2}$ & $79.3$ & $\mathbf{63.8}$ & $\mathbf{51.3}$ & $38.4$ \\ 
     \hline
   \end{tabular} 
\caption{Zero-shot learning results on SUN, CUB, AWA1, AWA2 and aPY using SS = Standard Split, PS = Proposed Split Version 2.0 with ResNet features. The results report top-1 accuracy in \%.   }
\label{tab:zeroshot}
\end{table*}
}

\begin{figure*}[t]
	\centering 
        \includegraphics[width=0.16\linewidth, angle=-90, trim=0 20 0 60,clip]{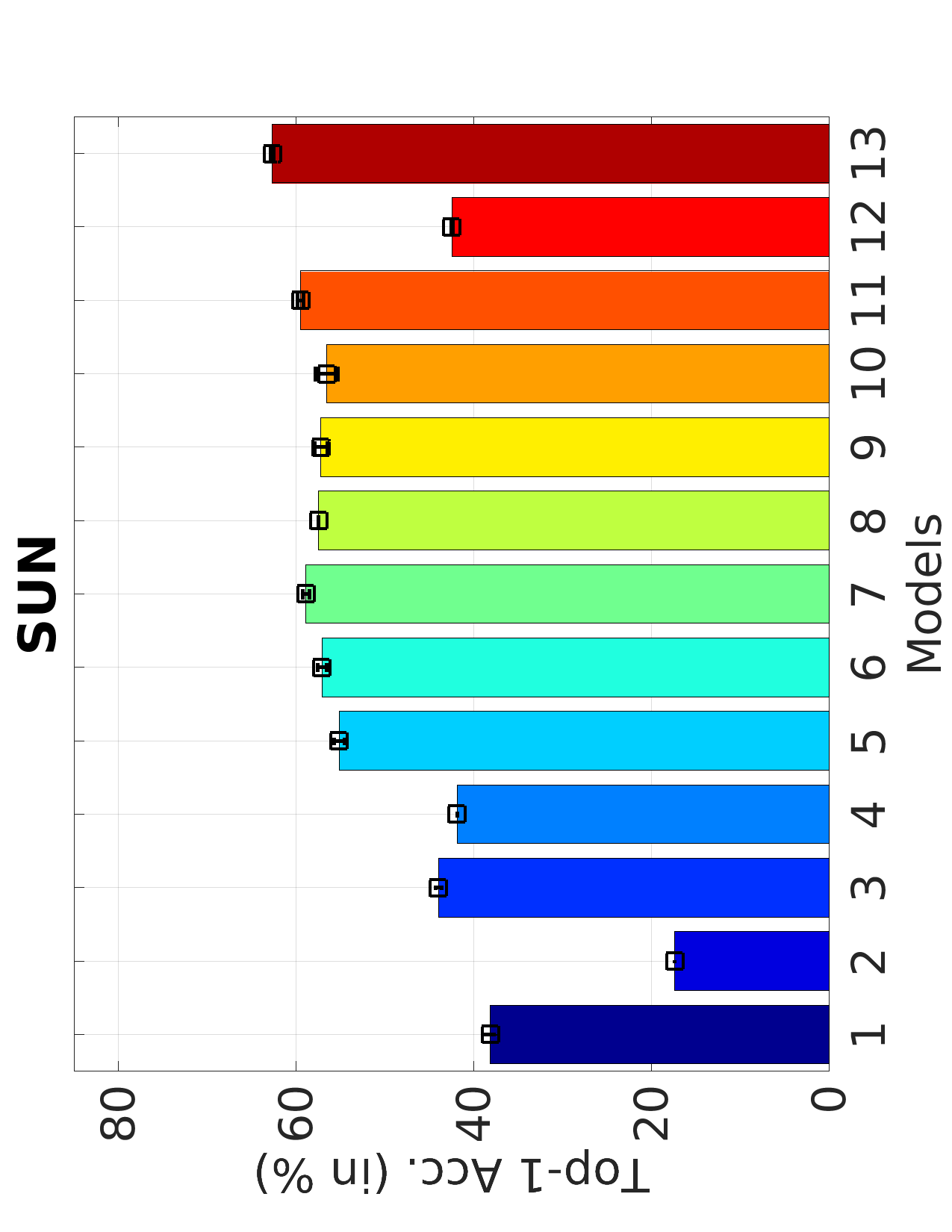}
    \includegraphics[width=0.16\linewidth, angle=-90, trim=0 20 0 60,clip]{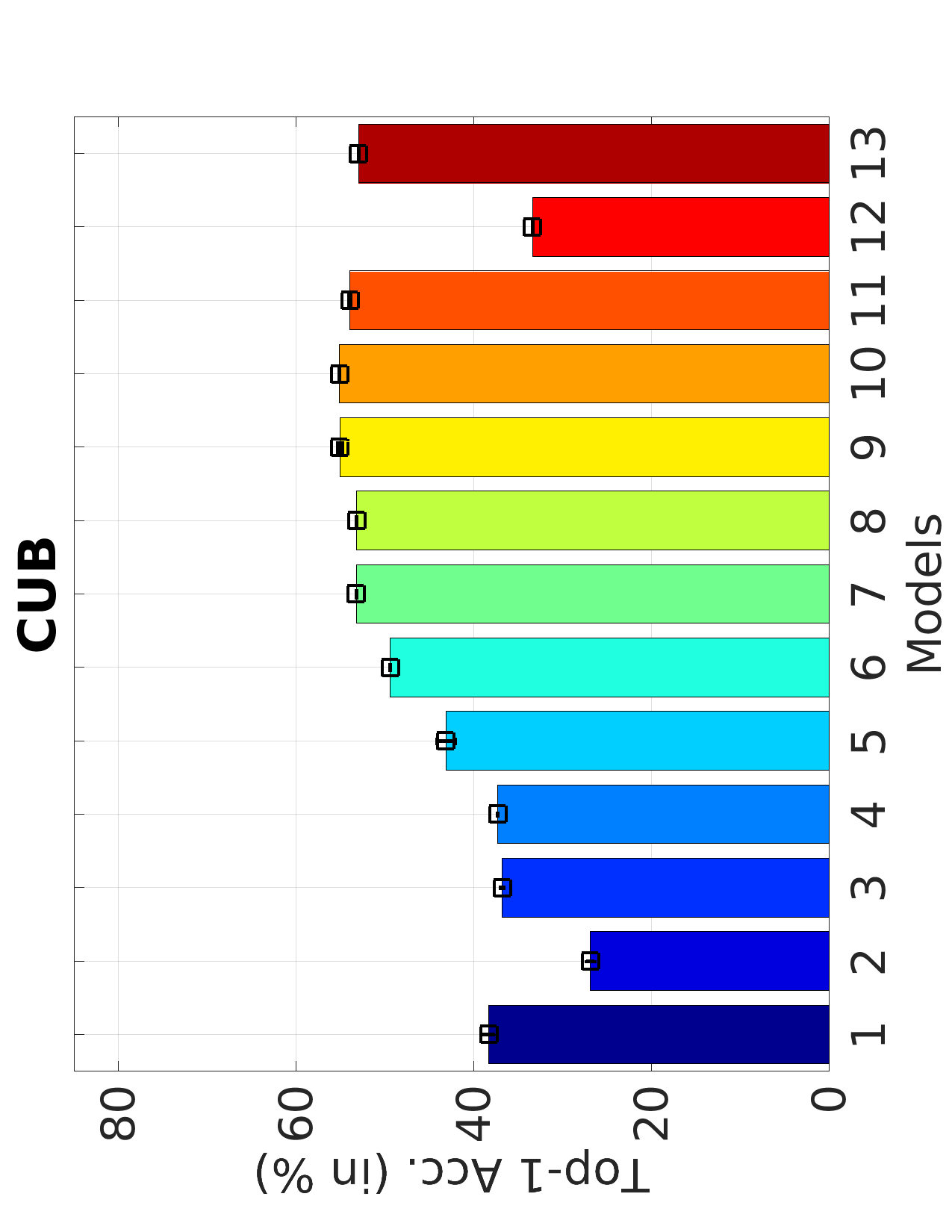}
    \includegraphics[width=0.16\linewidth, angle=-90, trim=0 20 0 60,clip]{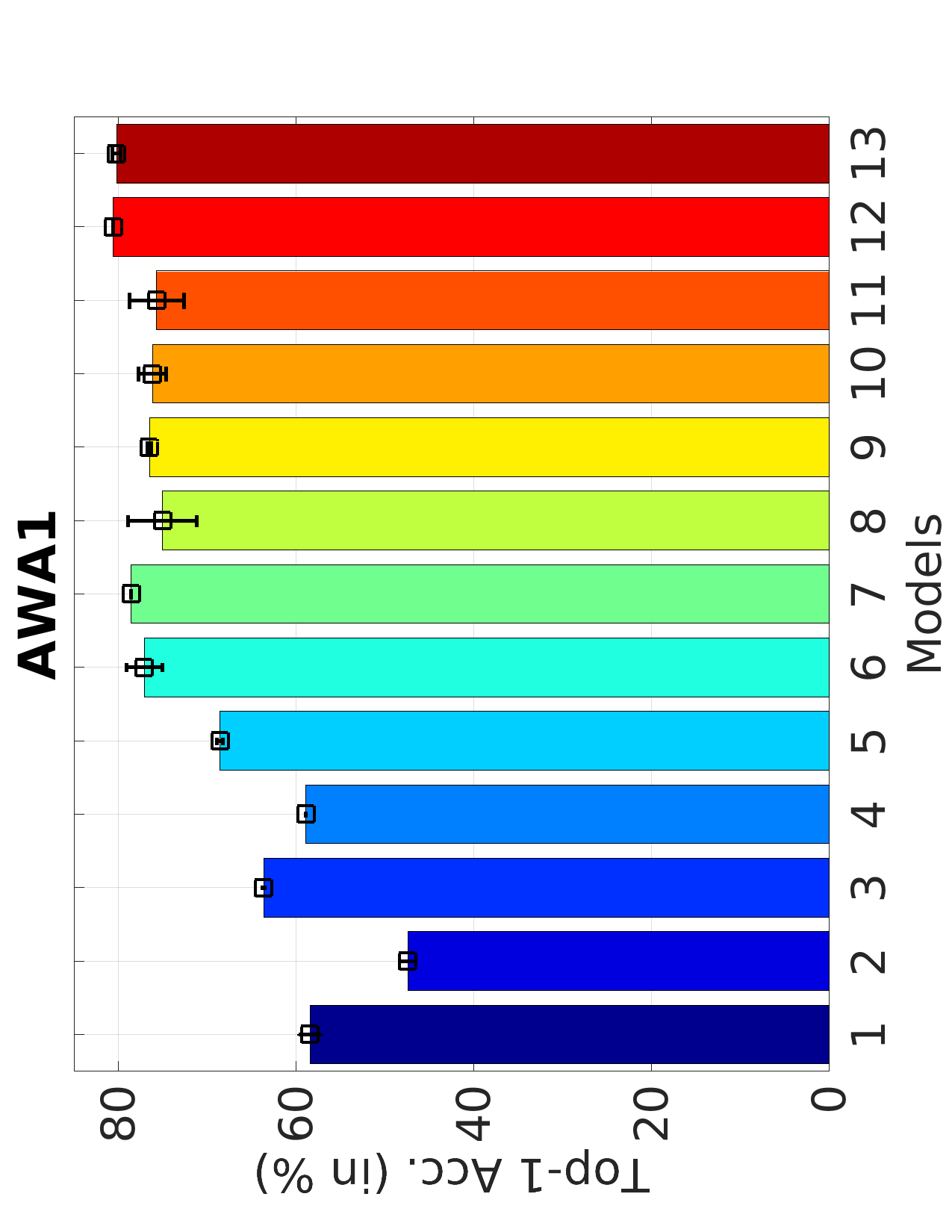}
    \includegraphics[width=0.16\linewidth, angle=-90, trim=0 20 0 60,clip]{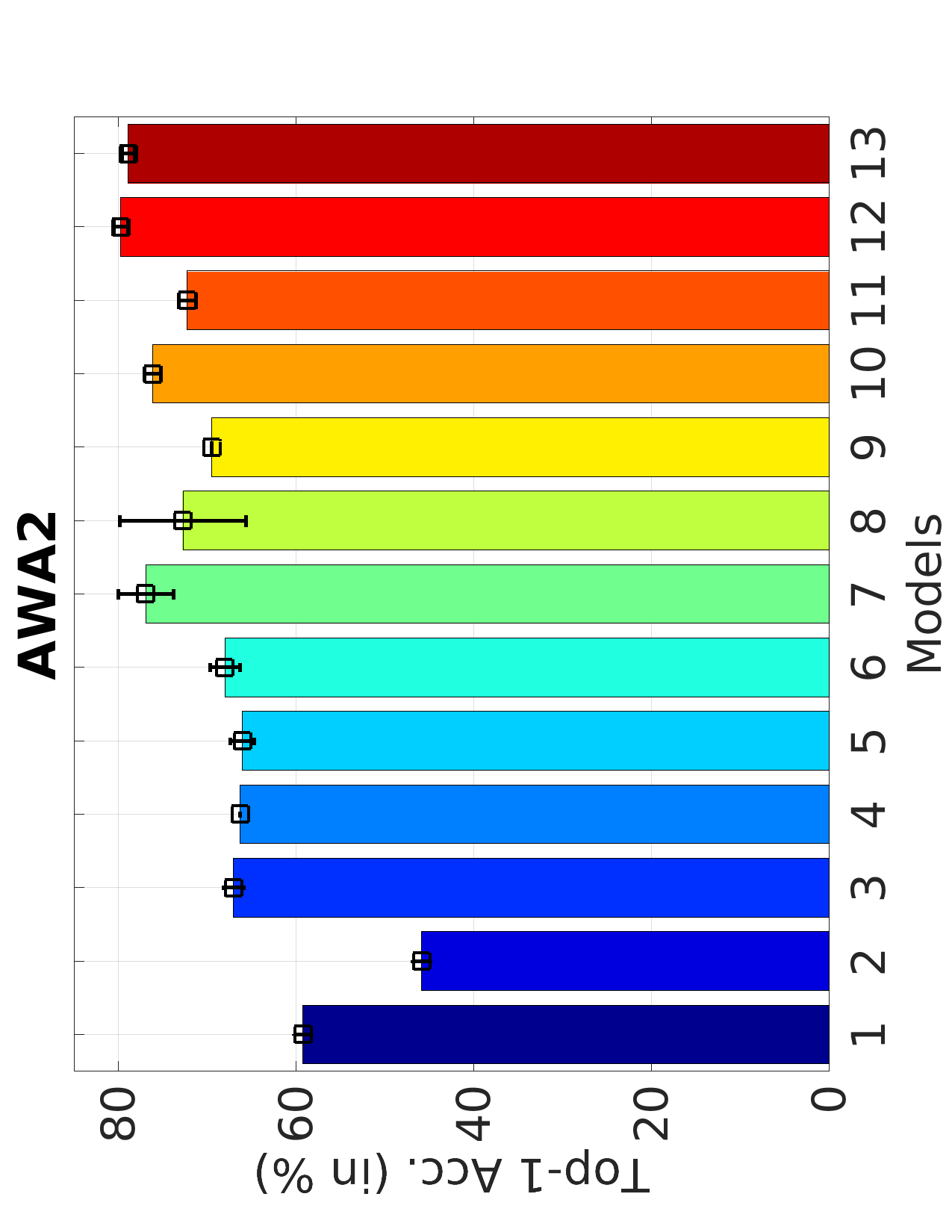}
    \includegraphics[width=0.16\linewidth, angle=-90, trim=0 20 0 60,clip]{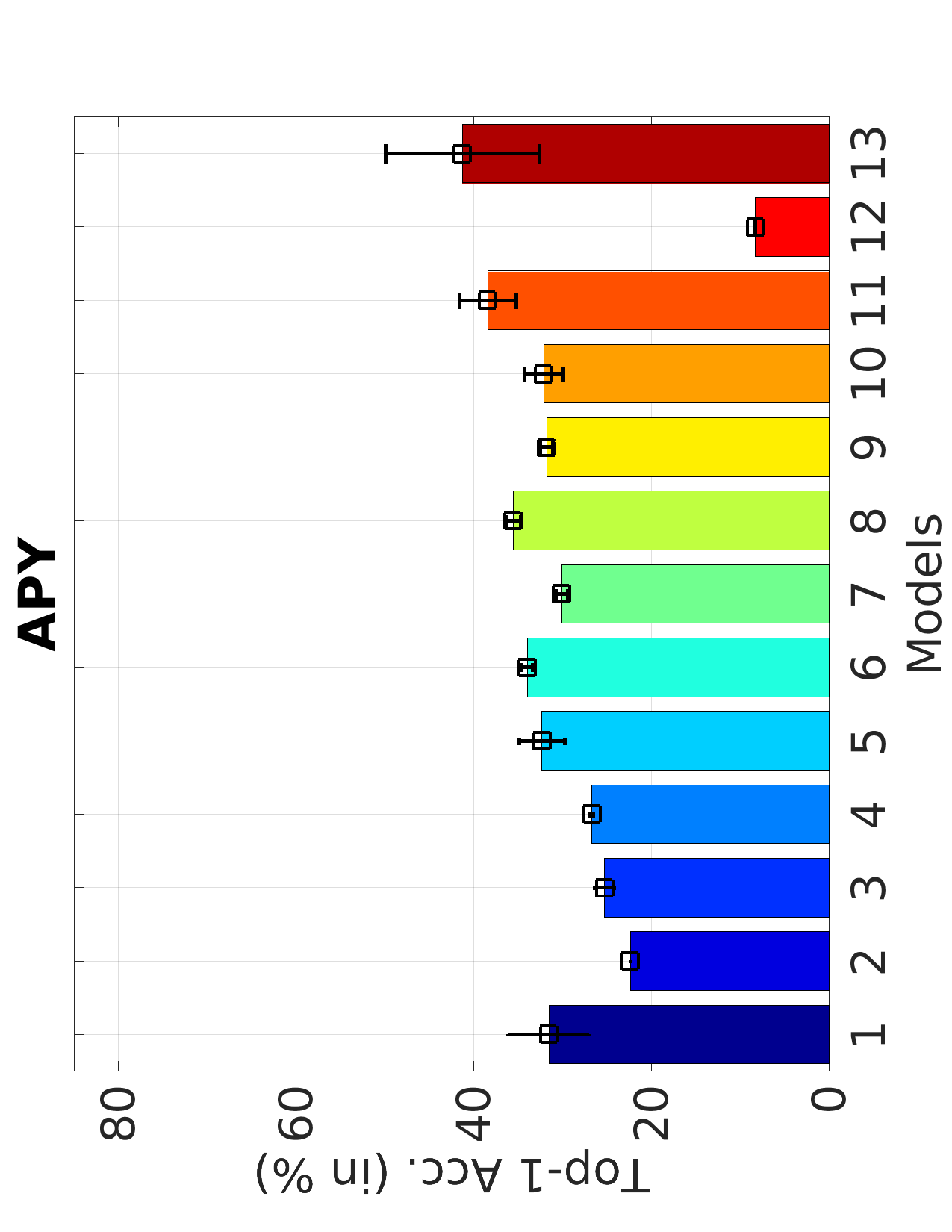} 
    \includegraphics[width=0.16\linewidth, angle=-90, trim=30 570 0 60,clip]{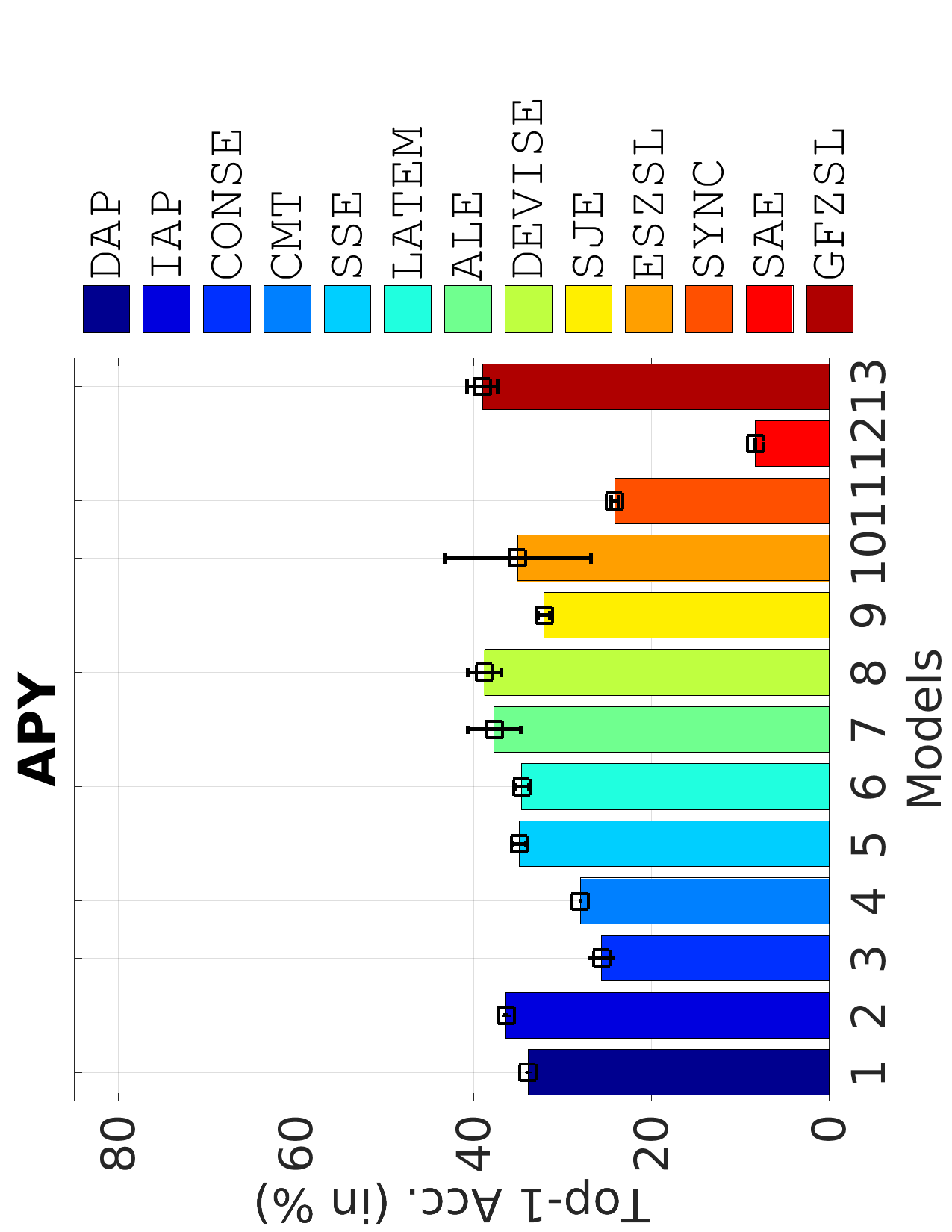}  \\
    \vspace{2mm}
    \includegraphics[width=0.16\linewidth, angle=-90, trim=0 20 0 60,clip]{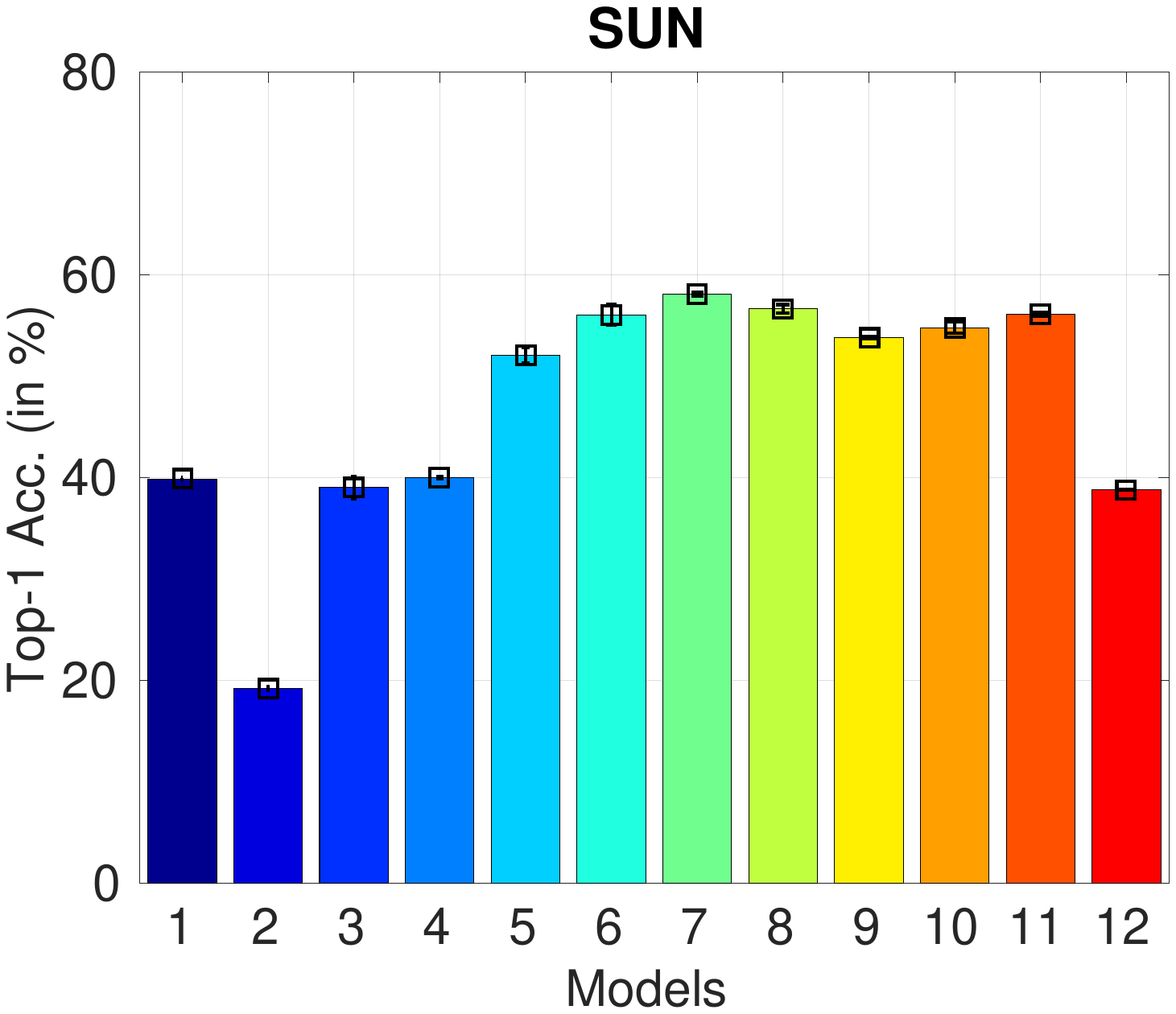}
	\includegraphics[width=0.16\linewidth, angle=-90, trim=0 20 0 60,clip]{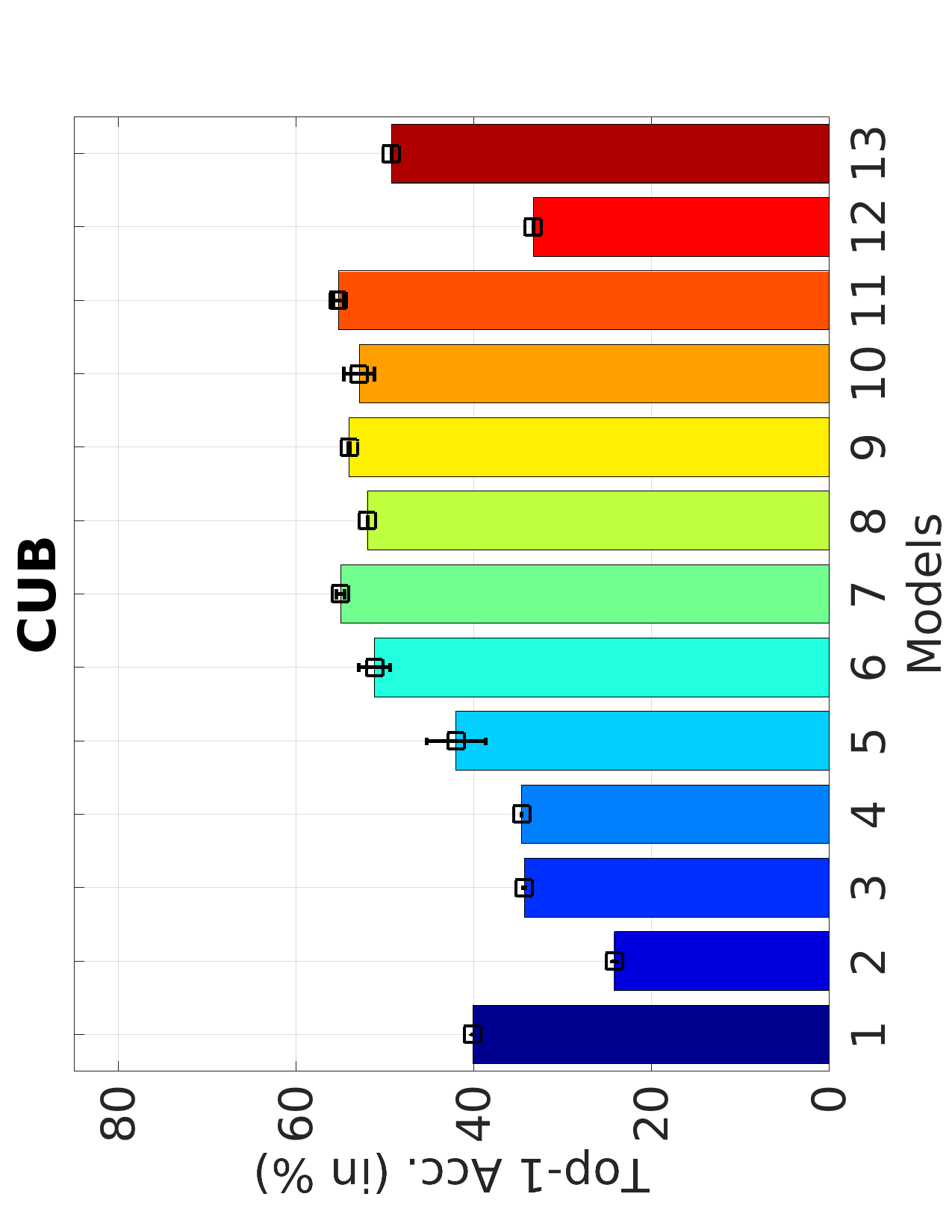}   
    \includegraphics[width=0.16\linewidth, angle=-90, trim=0 20 0 60,clip]{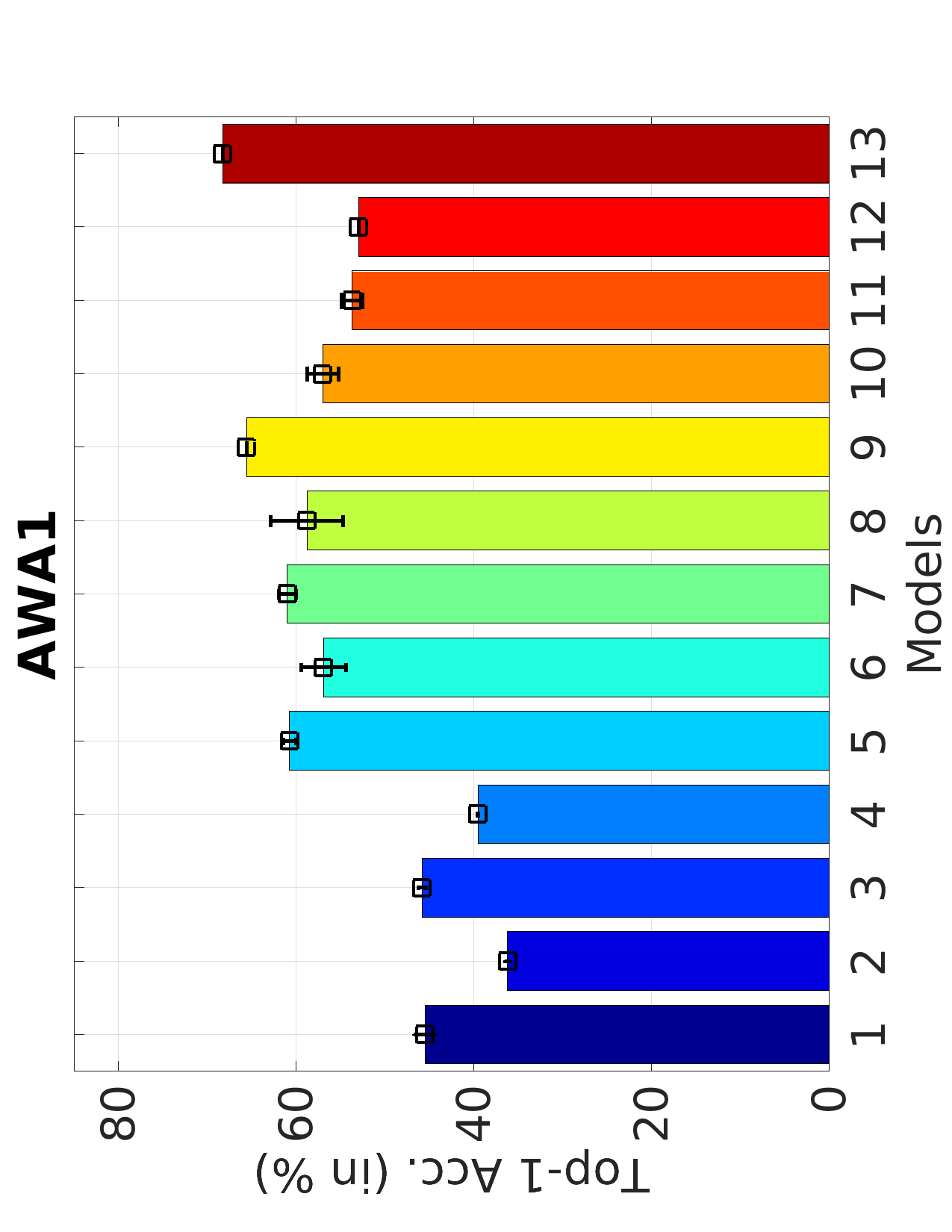}
    \includegraphics[width=0.16\linewidth, angle=-90, trim=0 20 0 60,clip]{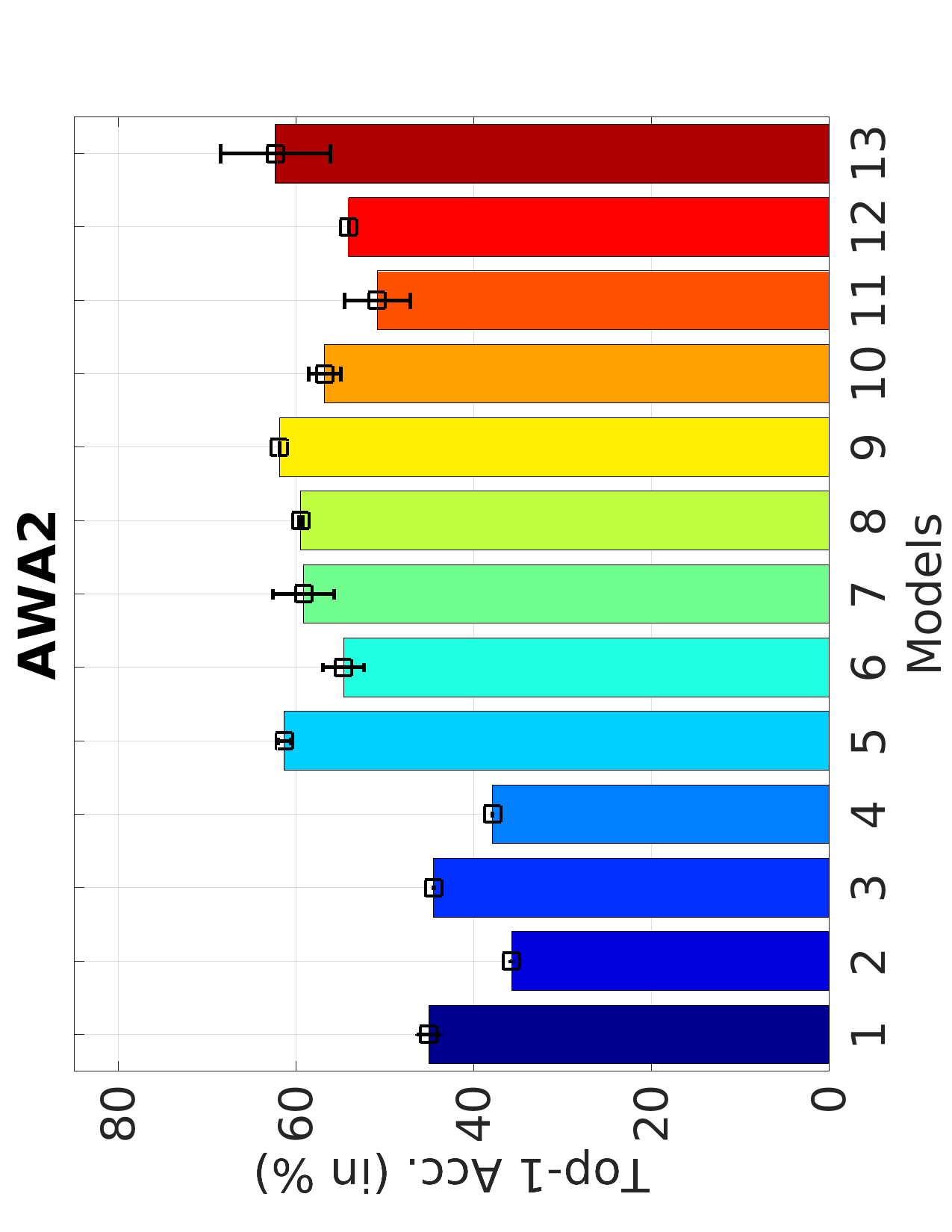}
    \includegraphics[width=0.16\linewidth, angle=-90, trim=0 20 0 60,clip]{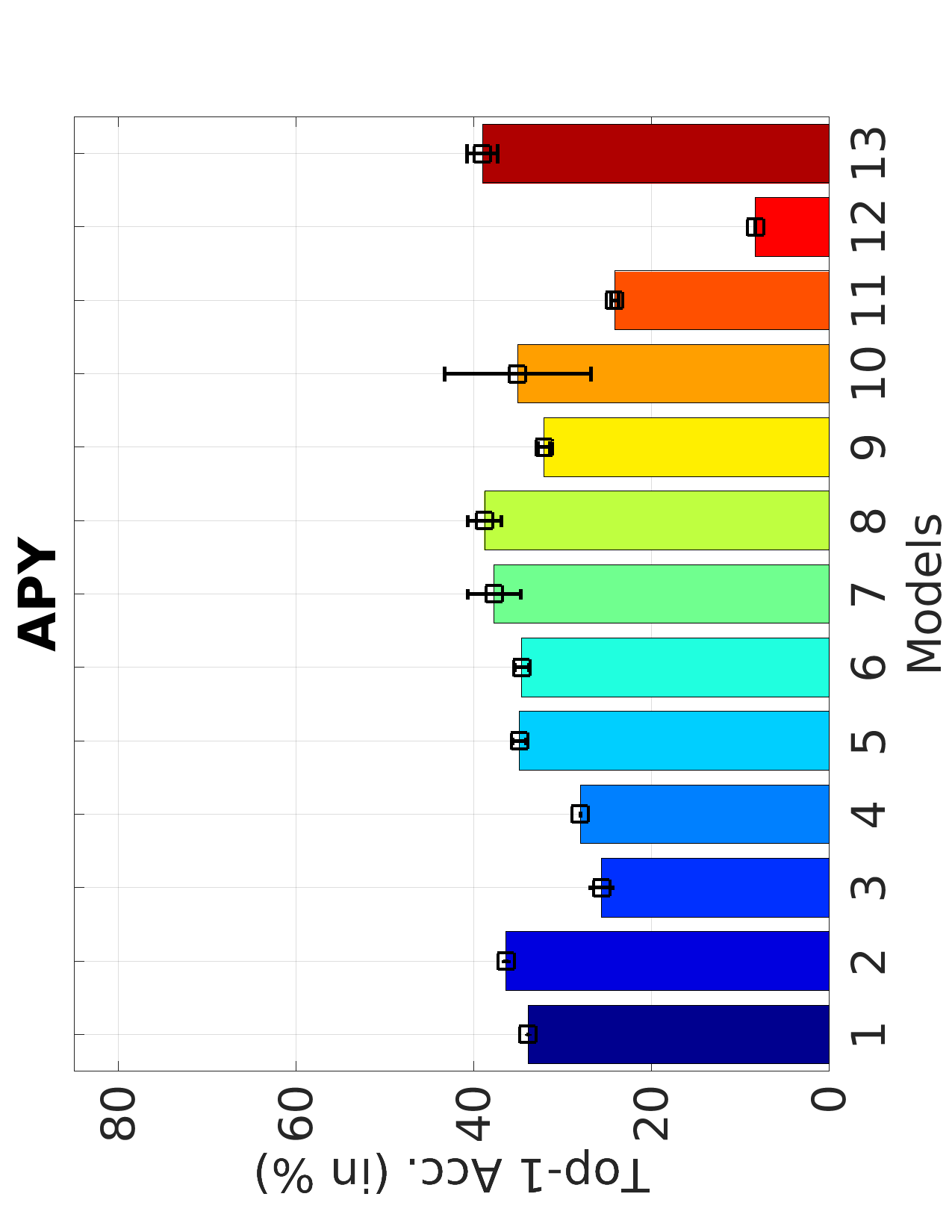}   
    \includegraphics[width=0.16\linewidth, angle=-90, trim=30 570 0 60,clip]{apy_legend}
    \caption{Robustness of 10 methods evaluated on SUN, CUB, AWA1, aPY using 3 validation set splits (results are on the same test split). Top: original split, Bottom: proposed split (Image embeddings = ResNet). We measure top-1 accuracy in \%.} 
	\label{fig:bar_plot}
\end{figure*}

\section{Experiments}
\label{sec:exp}
We first provide ZSL results on the attribute datasets SUN, CUB, AWA1, AWA2 and aPY and then on the large-scale ImageNet dataset. Finally, we present results for the GZSL setting.

\subsection{Zero-Shot Learning Experiments}
On attribute datasets, i.e. SUN, CUB, AWA1, AWA2, and aPY, we first reproduce the results of each method using their evaluation protocol, then provide a unified evaluation protocol using the same train/val/test class splits, followed by our proposed train/val/test class splits on SUN, CUB, AWA1, aPY and AWA2. We also evaluate the robustness of the methods to parameter tuning and visualize the ranking of different methods. Finally, we evaluate the methods on the large-scale ImageNet dataset.

\myparagraph{Comparing State-of-The-Art Models.} For sanity-check, we re-evaluate methods \cite{LNH13,ZV15,XASNHS16,ARWLS15,RT15,CCGS16} and \cite{kodirov2017semantic} using publicly available features and code from the original publication on SUN, CUB, AWA1 and aPY (CMT~\cite{SGMN13} evaluates on CIFAR dataset.). We observe from the results in~\autoref{tab:reproduce} that our reproduced results of DAP\cite{LNH13}, SYNC~\cite{CCGS16}, GFZSL~\cite{VR17}, GFZSL-tran~\cite{VR17}, DSRL~\cite{YG17} and SAE~\cite{kodirov2017semantic} are nearly identical to the reported number in their original publications. For LATEM~\cite{XASNHS16}, we obtain slightly different results which can be explained by the non-convexity and thus the sensibility to initialization. Similarly for SJE~\cite{ARWLS15} random sampling in SGD might lead to slightly different results. ESZSL~\cite{RT15} has some variance because its algorithm randomly picks a validation set during each run, which leads to different hyperparameters. Notable observations on SSE~\cite{ZV15} results are as follows. The published code has hard-coded hyperparameters operational on aPY, i.e. number of iterations, number of data points to train SVM, and one regularizer parameter $\gamma$ which lead to inferior results than the ones reported here, therefore we set these parameters on validation sets. On SUN, SSE uses $10$ classes (instead of $72$) and our results with validated parameters got an improvement of $0.5\%$ that may be due to random sampling of training images. On AWA1, our reproduced result being $64.9\%$ is significantly lower than the reported result ($76.3\%$). However, we could not reach the reported result even by tuning parameters on the test set ($73.8\%$).

In addition to \cite{LNH13,ZV15,XASNHS16,ARWLS15,RT15,CCGS16,SGMN13,kodirov2017semantic}, we re-implement~\cite{NMBSSFCD14,FCSBDRM13,APHS15} based on the original publications. We use train, validation, test splits as provided in~\autoref{tab:datasets} and report results in~\autoref{tab:zeroshot} with deep ResNet features. DAP~\cite{LNH13} uses hand-crafted image features and thus reproduced results with those features are significantly lower than the results with deep features ($22.1\%$ vs $38.9\%$). When we investigate the results in detail, we noticed two irregularities with reported results on SUN. First, SSE~\cite{ZV15} and ESZSL~\cite{RT15} report results on a test split with $10$ classes whereas the standard split of SUN contains $72$ test classes ($74.5\%$ vs $54.5\%$ with SSE~\cite{ZV15} and $64.3\%$ vs $57.3\%$ with ESZSL~\cite{RT15}). Second, after careful examination and correspondence with the authors of SYNC~\cite{CCGS16}, we detected that SUN features were extracted with a MIT Places~\cite{zhou2014learning} pre-trained model. As the MIT Places dataset intersects with both training and test classes of SUN, it is expected to lead to significantly better results than ImageNet pre-trained models ($62.8\%$ vs $59.1\%$). In addition, while SAE~\cite{kodirov2017semantic} reported $84.7\%$ on AWA1, we obtain only $80.7\%$ on the standard split. This could be explained by two differences. First, we measure per-class accuracy but SAE~\cite{kodirov2017semantic} reports per-image accuracy which is typically higher when the dataset is class-imbalanced, e.g. AWA1. Indeed, their reported accuracy decreases to $82.0\%$ if per-class accuracy is applied. Second, we confirmed with the authors of SAE~\cite{kodirov2017semantic} that they improved GoogleNet~\cite{szegedy2014going} by adding Batch Normalization and averaging 5 randomly cropped images to obtain better image features. Therefore, as expected, improving visual features lead to improved results in zero-shot learning. 

\myparagraph{Promoting Our Proposed Splits (PS).} We propose new dataset splits (see details in~\autoref{sec:datasets}) ensuring that test classes of any of the datasets do not overlap with the ImageNet1K used to pre-train ResNet. As training ResNet is a part of the training procedure, including test classes in the dataset used for pre-training ResNet would violate the zero-shot learning conditions. We compare the results obtained with our proposed split (PS) with previously published standard split (SS) results in~\autoref{tab:zeroshot}.

Our first observation is that the results on the PS are significantly lower than the SS for AWA1 and AWA2. This is expected as most of the test classes of AWA1 and AWA2 in SS overlaps with ImageNet 1K. On the other hand, for fine-grained datasets CUB and SUN, the results are not significantly effected as the overlap in that case was not as significant.  Our second observation regarding the method ranking is as follows. On SS, GFZSL~\cite{VR17} is the best performing method on SUN ($62.9\%$) and aPY ($51.3\%$) datasets whereas SJE~\cite{ARWLS15} performs the best on CUB ($55.3\%$) and SAE~\cite{kodirov2017semantic} performs the best on AWA1 ($80.6\%$) and AWA2 ($80.7\%$) dataset. On PS, GFZSL~\cite{VR17} performs the best on SUN ($60.6\%$), AWA1 ($68.2\%$) and AWA2 ($63.8\%$), ALE~\cite{APHS15} on aPY ($39.7\%$), and SYNC~\cite{CCGS16} on CUB ($56.0\%$). ALE, SJE and DEVISE all use max-margin bi-linear compatibility learning framework which seem to perform better than others. 

\myparagraph{Evaluating Robustness.} We evaluate robustness of $13$ methods, i.e. \cite{LNH13,ZV15,XASNHS16,ARWLS15,RT15,CCGS16,SGMN13,NMBSSFCD14,FCSBDRM13,APHS15,kodirov2017semantic,VR17}, to hyperparameters by setting them on $3$ different validation splits while keeping the test split intact. We report results on SS (\autoref{fig:bar_plot}, top) and PS (\autoref{fig:bar_plot}, bottom) for SUN, CUB, AWA1, AWA2 and aPY datasets. On SUN and CUB, the results are stable across methods and across dataset splits. This is expected as these datasets both have a balanced number of images across classes and they are fine-grained datasets. Therefore, the validation splits are similar. On the other hand, aPY being a small and coarse-grained dataset has several issues. First, many of the test classes of aPY are included in ImageNet1K. Second, it is not well balanced, i.e. different validation class splits contain significantly different number of images. Third, the class embeddings are far from each other, i.e. objects are semantically different, therefore different validation splits learn a different mapping between images and classes. On AWA1 and AWA2, on SS, the DEVISE method seems to show the largest variance. This might be due to the fact that AWA1 and AWA2 datasets are also coarse-grained and test classes overlap with ImageNet training classes. Indeed, AWA2 being slightly more balanced than AWA1, in the proposed split it does not lead to such a high variance for DEVISE.

\begin{figure}[t]
	\centering
	\includegraphics[width=0.48\columnwidth, trim=0 60 70 0,clip]{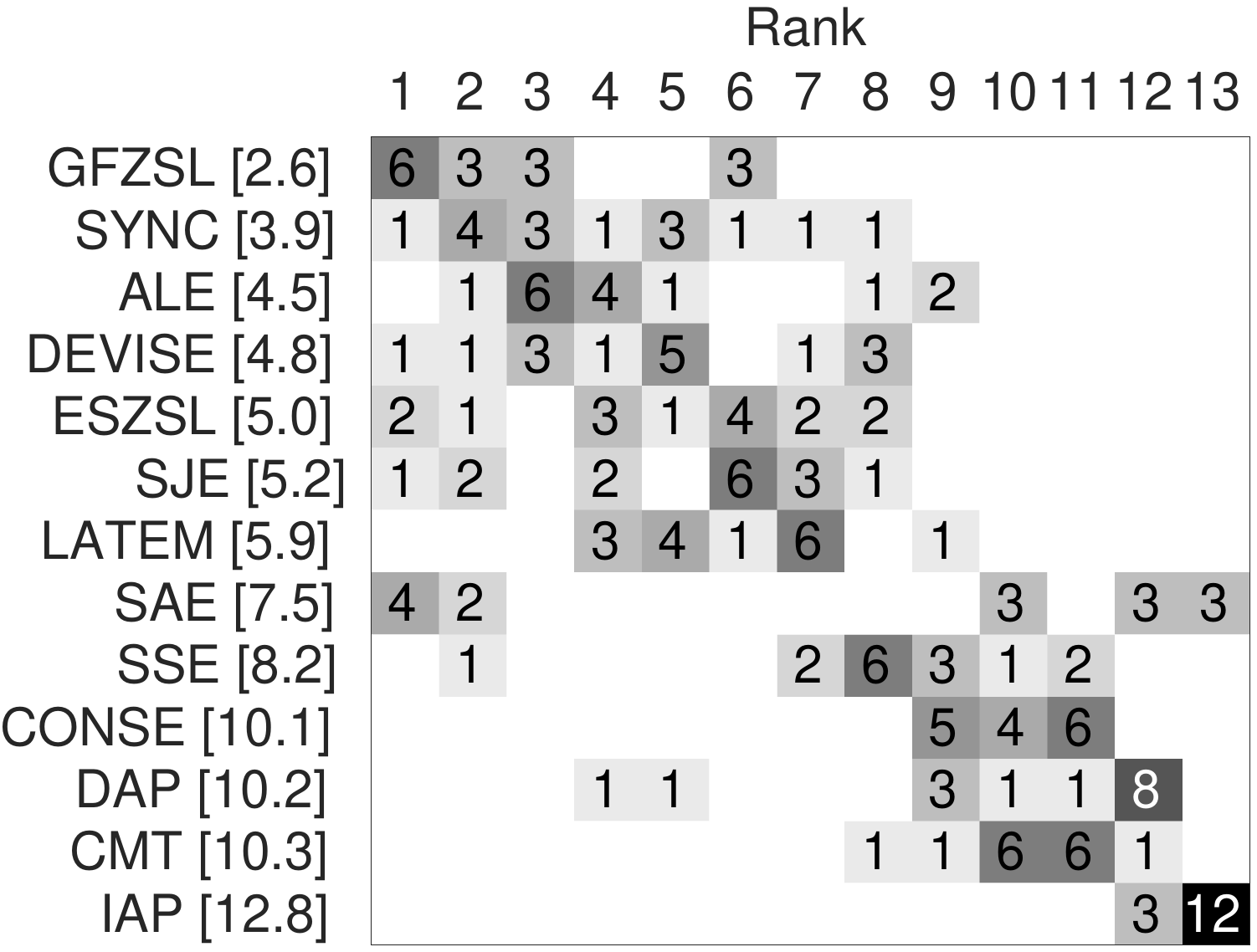}
    \includegraphics[width=0.48\columnwidth, trim=0 60 70 0,clip]{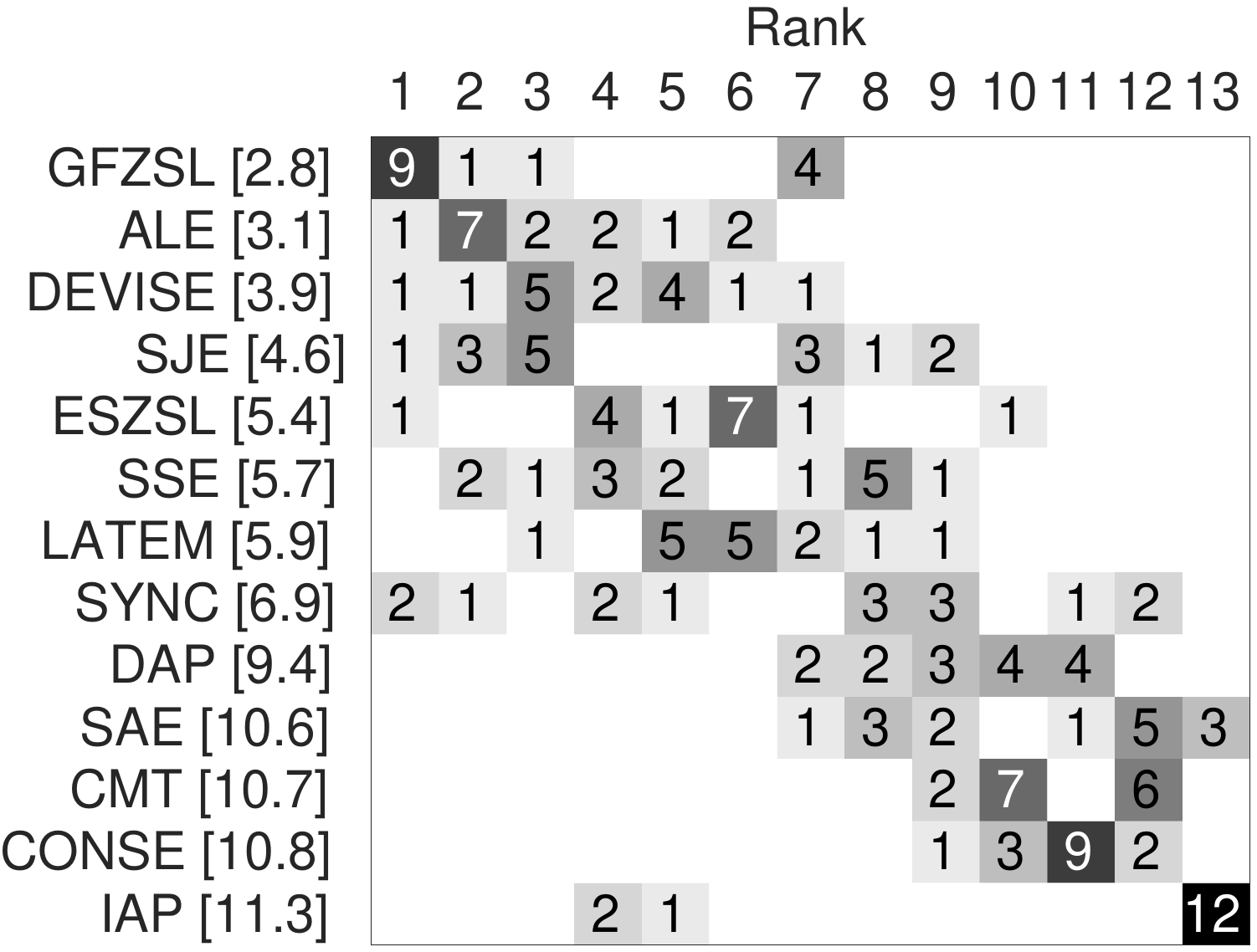}
    \caption{Ranking 12 models by setting parameters on three validation splits on the standard split (SS, left) and proposed split Version 2.0 (PS, right) setting. Element $(i, j)$ indicates number of times model $i$ ranks at $j$th over all $4\times3$ observations. Models are ordered by their mean rank (displayed in brackets).} \vspace{-3mm}
	\label{fig:rank_matrix}
\end{figure}

\myparagraph{Visualizing Method Ranking.} We first evaluate the $13$ methods using three different validation splits as in the previous experiment. We then rank them based on their per-class top-1 accuracy using the non-parametric Friedman test~\cite{GH08}, which does not assume a distribution on performance but rather uses algorithm ranking. Each entry of the rank matrix on~\autoref{fig:rank_matrix} indicates the number of times the method is ranked at the first to thirteenth rank. We then compute the mean rank of each method and order them based on the mean rank across datasets. 

Our general observation is that the highest ranked method on both splits is GFZSL, the second highest ranked method on the standard split (SS) is SYNC while it drops to the seventh rank on the proposed split (PS). On the other hand, ALE ranks the second on the SS and the first on the PS. We reinforce our initial observation from numerical results and conclude that GFZSL and ALE seems to be the method that is the most robust in zero-shot learning setting for attribute datasets. These results also indicate the importance of choosing zero-shot splits carefully. On the PS, the two of three highest ranked methods are compatibility learning methods, i.e. ALE and DEVISE whereas the three lowest ranked methods are attribute classifier learning or hybrid methods, i.e. IAP, CMT and CONSE. Therefore, max-margin compatibility learning methods lead to consistently better results in the zero-shot learning task compared to learning independent classifiers. Finally, visualizing the method ranking in this way provides a visually interpretable way of how models compare across datasets.

{
\setlength{\tabcolsep}{3pt}
\renewcommand{\arraystretch}{1}
\begin{table}[t]
 \centering
      \begin{tabular}{l c c c c}
     & \multicolumn{4}{c}{\textbf{Training Set : Test Set}} \\
\textbf{Method} & \textbf{AWA1:AWA1} & \textbf{AWA1:AWA2} & \textbf{AWA2:AWA2} & \textbf{AWA2:AWA1}   \\    
     \hline
     DAP~\cite{LNH13} & $44.1$ & $44.2$ & $46.1$ & $46.2$ \\
     IAP~\cite{LNH13} & $35.9$ & $36.1$ & $35.9$ & $35.3$ \\
     CONSE~\cite{NMBSSFCD14} & $45.6$ & $46.5$ & $44.5$ & $43.7$ \\
     CMT~\cite{SGMN13} & $39.5$ & $40.7$ & $37.9$ & $37.7$ \\
     SSE~\cite{ZV15} & $60.1$ & $61.6$ & $61.0$ & $59.8$ \\
     LATEM~\cite{XASNHS16} & $55.1$ & $55.4$ & $55.8$ & $53.5$ \\
     ALE~\cite{APHS15} & $59.9$ & $59.9$ & $62.5$ & $60.9$ \\
     DEVISE~\cite{FCSBDRM13} & $54.2$ & $55.2$ & $59.7$ & $57.7$ \\
     SJE~\cite{ARWLS15} & $65.6$ & $65.5$ & $61.9$ & $62.0$ \\
     ESZSL~\cite{RT15} & $58.2$ & $58.5$ & $58.6$ & $59.9$ \\
     SYNC~\cite{CCGS16} & $54.0$ & $53.7$ & $46.6$ & $46.9$\\ 
     SAE~\cite{kodirov2017semantic} & $53.0$ & $52.4$ & $54.1$ & $53.1$ \\ 
     \hline
   \end{tabular} 
\caption{Cross-dataset evaluation over AWA1 and AWA2 in zero-shot learning setting on the Proposed Splits: Left of the colon indicates the training set and right of the colon indicates the test set, e.g. AWA1:AWA2 means that the model is trained on the train set of AWA1 and evaluated on the test set of AWA2. We measure top-1 accuracy in \%. }
\label{tab:awa_awa2}
\end{table}
}

{
\renewcommand{\arraystretch}{1}
\begin{table*}[t]
 \centering
   \begin{tabular}{l c c |c c c |c c c |c}
     & \multicolumn{2}{c}{\textbf{Hierarchy}} & \multicolumn{3}{c}{\textbf{Most Populated}} & \multicolumn{3}{c}{\textbf{Least Populated}} & \textbf{All} \\
     \textbf{Method} & \textbf{2 H} & \textbf{3 H} & \textbf{500} & \textbf{1K} & \textbf{5K} & \textbf{500} & \textbf{1K} & \textbf{5K} & \textbf{20K} \\     
     \hline
     CONSE~\cite{NMBSSFCD14} & $7.63$ & $2.18$ & $12.33$ & $8.31$ & $3.22$ & $3.53$ & $2.69$ & $1.05$ & $0.95$ \\
     CMT~\cite{SGMN13} & $2.88$ & $0.67$ & $5.10$ & $3.04$ & $1.04$ & $1.87$ & $1.08$ & $0.33$ & $0.29$ \\
     LATEM~\cite{XASNHS16} & $5.45$ & $1.32$ & $10.81$ & $6.63$ & $1.90$ & $4.53$ & $2.74$ & $0.76$ & $0.50$ \\
      ALE~\cite{APHS15} & $5.38$ & $1.32$ & $10.40$ & $6.77$ & $2.00$ & $4.27$ & $2.85$ & $0.79$& $0.50$ \\
     DEVISE~\cite{FCSBDRM13} & $5.25$ & $1.29$ & $10.36$ & $6.68$ & $1.94$ & $4.23$ & $2.86$ & $0.78$ & $0.49$ \\
     SJE~\cite{ARWLS15} & $5.31$ & $1.33$ & $9.88$ & $6.53$ & $1.99$ & $4.93$ & $2.93$ & $0.78$ & $0.52$ \\   
     ESZSL~\cite{RT15} & $6.35$ & $1.51$ & $11.91$ & $7.69$ & $2.34$ & $4.50$ & $3.23$ & $0.94$ & $0.62$ \\
     SYNC~\cite{CCGS16} & $\mathbf{9.26}$ & $\mathbf{2.29}$ & $\mathbf{15.83}$ & $\mathbf{10.75}$ & $\mathbf{3.42}$ & $\mathbf{5.83}$ & $\mathbf{3.52}$ & $\mathbf{1.26}$ & $\mathbf{0.96}$ \\
     SAE~\cite{kodirov2017semantic} & $4.89$ & $1.26$ & $9.96$ & $6.57$ & $2.09$ & $2.50$ & $2.17$ & $0.72$ & $0.56$ \\
     GFZSL~\cite{VR17} & $1.45$ & $--$ & $2.01$ & $1.35$ & $--$ & $1.40$ & $1.11$ & $0.13$ & $--$ \\
     \hline
   \end{tabular} 
\caption{ImageNet with different splits: 2/3 H = classes with 2/3 hops away from the $\mathcal{Y}^{tr}$ of ImageNet1K, 500/1K/5K most populated classes, 500/1K/5K least populated classes, All = The remaining 20K categories of ImageNet ($\mathcal{Y}^{ts}$). We measure top-1 accuracy in \%.}
\label{tab:ImageNet}
\end{table*}
}

\begin{figure*}[t]
	\centering
        \includegraphics[width=0.28\linewidth,angle=-90, trim=0 30 0 70,clip]{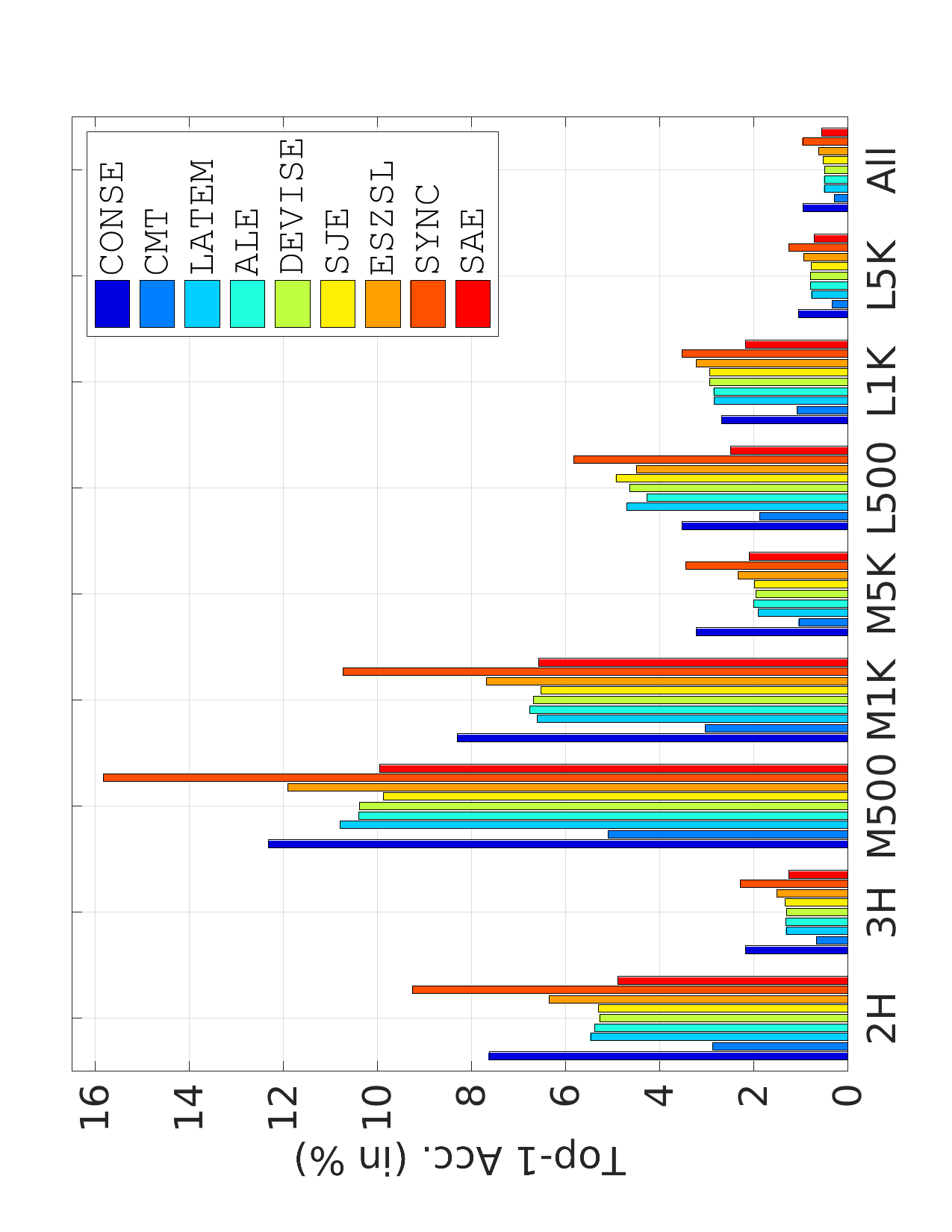}
        \includegraphics[width=0.28\linewidth, angle=-90, trim=0 30 0 70,clip]{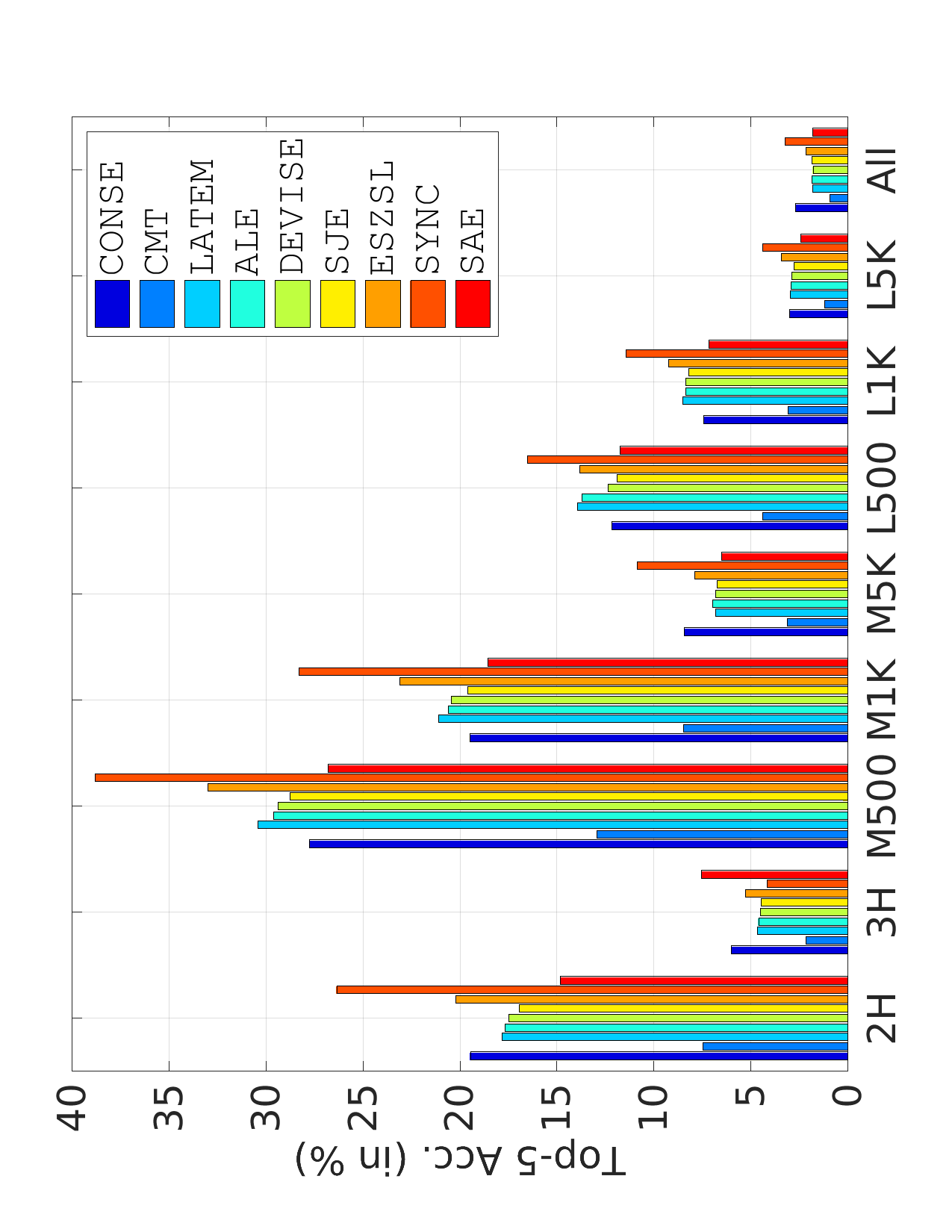}
        \includegraphics[width=0.28\linewidth, angle=-90, trim=0 30 0 70,clip]{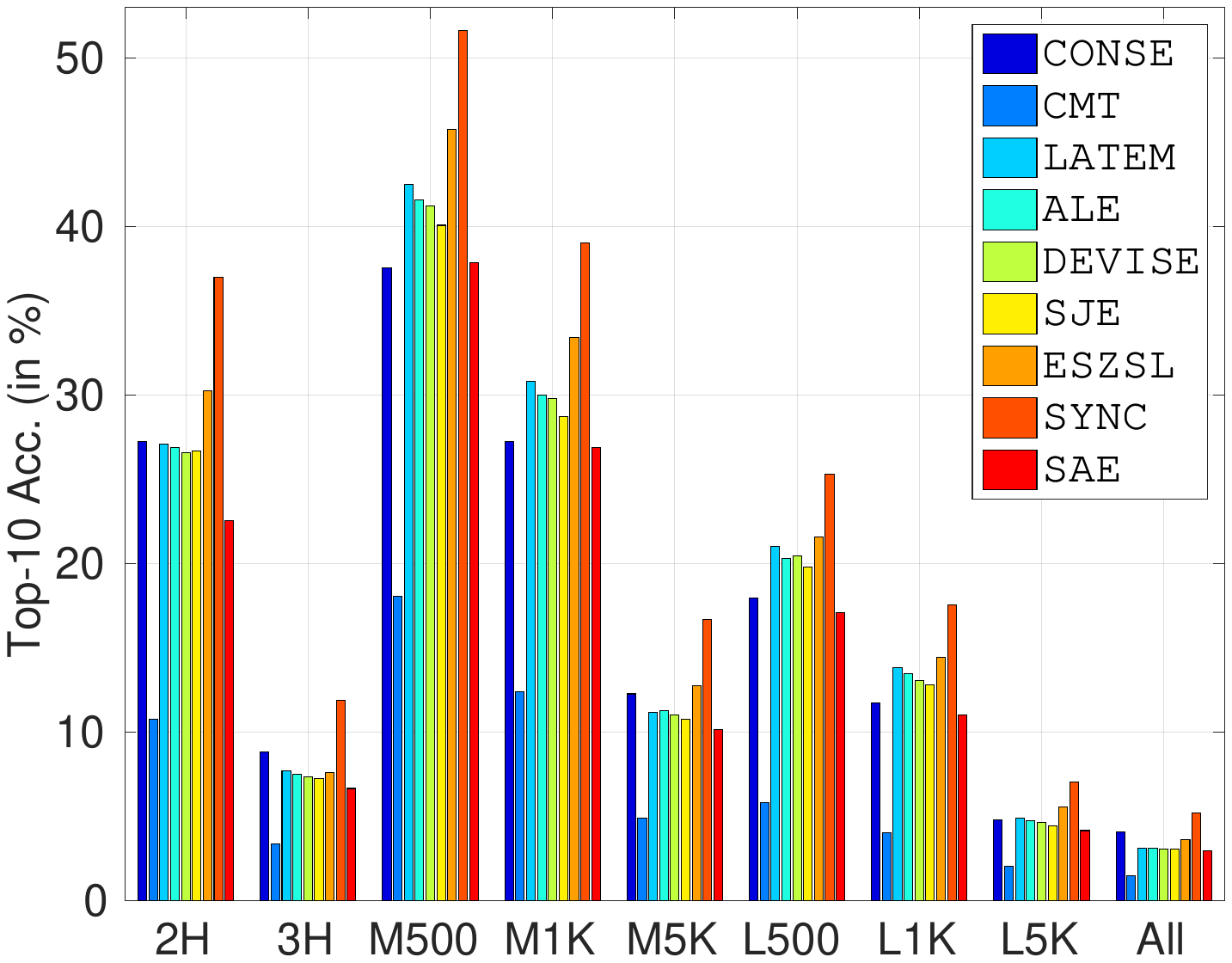}
	\caption{Zero-Shot Learning experiments on Imagenet, measuring Top-1, Top-5 and Top-10 accuracy. 2/3 H = classes with 2/3 hops away from ImageNet1K training classes ($\mathcal{Y}^{tr}$), M500/M1K/M5K denote 500, 1K and 5K most populated classes, L500/L1K/L5K denote 500, 1K and 5K least populated classes, All = The remaining 20K categories of ImageNet.}
	\label{fig:imagenet_zsl_top1}
\end{figure*}

\myparagraph{Results on Our Proposed AWA2.} We introduce AWA2 which has the same classes and attributes as AWA1, but contains different images each coming with a public copyright license. In order to show that AWA1 and AWA2 images are not the same but similar in nature, we compare the zero-shot learning results on AWA1 and AWA2 in Table.~\ref{tab:zeroshot}. Under the Standard Splits (SS), SAE~\cite{kodirov2017semantic} is the best performing method on both AWA1~($80.6\%$) and AWA2~($80.7\%$). Similarly, for most of the methods, the results on AWA1 are close to those on AWA2, for instance, DAP obtains $57.1\%$ on AWA1 and $58.7\%$ on AWA2, SSE obtains $68.8\%$ on AWA1 and $67.5\%$ AWA2, etc. The results under the Proposed Splits (PS) are also consistent across AWA1 and AWA2. For $8$ out of $12$ methods, the performance difference between AWA1 and AWA2 is within $2\%$. On the other hand, the same consistency is not observed for DEVISE~\cite{FCSBDRM13}, SJE~\cite{ARWLS15} and SYNC~\cite{CCGS16}. For instance, SJE~\cite{ARWLS15} obtains $65.6\%$ on AWA1 and $61.9\%$ on AWA2. After careful examination, we noticed that SJE~\cite{ARWLS15} selects different hyperparameters for AWA1 and AWA2, which results in different performance on those two datasets. In our opinion, this does not indicate a possible dataset artifact, however shows that zero-shot learning is sensitive to parameter setting. 

Commonly, a model is trained and evaluated on the same dataset. Across dataset experiments are not easy as different datasets do not share the same attributes. However, AWA1 and AWA2 share both classes and attributes. In order to verify that AWA2 is a good replacement for AWA1, we conduct across-dataset evaluation for 12 methods, i.e. \cite{LNH13,ZV15,XASNHS16,ARWLS15,RT15,CCGS16,SGMN13,NMBSSFCD14,FCSBDRM13,APHS15,kodirov2017semantic}. In particular, with our Proposed Splits (PS), we train one model on the training set of AWA1 and evaluate it on the test set of AWA2 in the zero-shot learning setting, and vice versa. From Table.~\ref{tab:awa_awa2}, we observe that all the models trained on AWA1 generalize well to AWA2 and vice versa. 

In addition, we notice that the cross-dataset result is dependent on the training set. For instance, for all the methods, if we fix training set to be from AWA1, the results on the test set of AWA1 and AWA2 are close. To verify this hypothesis, we performed a paired t-test which determines if the mean difference between paired results is significantly higher than zero. To that end, we take the 24 pairs of results whose test sets are the same, i.e. the results obtained with 12 methods on AWA1:AWA2 and AWA2:AWA2 (2nd and 3rd column) as well as the results obtained with 12 methods on AWA1:AWA1 and AWA2:AWA1 (1st and 4th column). The paired t-test rejects the null hypothesis with p-value$=0.007$, indicating that the results are significantly different if the test set is the same but the training set is different. As a conclusion, the training set is an important indicator of the final result and the two datasets, i.e. AWA1 and AWA2 are sufficiently similar. Therefore, our cross-dataset experimental results indicate that AWA2 is a good replacement for AWA1.

{
\renewcommand{\arraystretch}{1}
\begin{table*}[t]
 \centering
   \begin{tabular}{l c c c |c c c |c c c |c c c | c c c}
      & \multicolumn{3}{c}{\textbf{SUN}} & \multicolumn{3}{c}{\textbf{CUB}} & \multicolumn{3}{c}{\textbf{AWA1}} & \multicolumn{3}{c}{\textbf{AWA2}} & \multicolumn{3}{c}{\textbf{aPY}}     \\
     \textbf{Method} &  \textbf{ts} & \textbf{tr} & \textbf{H} & \textbf{ts} & \textbf{tr} & \textbf{H} & \textbf{ts} & \textbf{tr} & \textbf{H} & \textbf{ts} & \textbf{tr} & \textbf{H} & \textbf{ts} & \textbf{tr} & \textbf{H}\\
     \hline
     DAP~\cite{LNH13} & $4.2$ &$25.1$ & $7.2$ & $1.7$ & $67.9$ & $3.3$ & $0.0$ & $88.7$ & $0.0$ & $0.0$ & $84.7$ & $0.0$ & $4.8$ & $78.3$ & $9.0$   \\
     IAP~\cite{LNH13} & $1.0$ &$37.8$ & $1.8$ & $0.2$ & $72.8$ & $0.4$ & $2.1$ & $78.2$ & $4.1$ & $0.9$ & $87.6$ & $1.8$ & $5.7$ & $65.6$ & $10.4$   \\
     CONSE~\cite{NMBSSFCD14} & $6.8$ & $35.9$ & $11.4$ & $2.0$ & $70.6$ & $3.9$ & $0.4$ & $\mathbf{89.6}$ & $0.8$ & $0.5$ & $\mathbf{90.6}$ & $1.0$ & $0.0$ & $\mathbf{91.2}$ & $0.0$   \\
     CMT~\cite{SGMN13} & $8.1$ & $21.8$ & $11.8$ & $7.2$ & $49.8$ & $12.6$ & $0.9$ & $87.6$ & $1.8$ & $0.5$ & $90.0$ & $1.0$ &$1.4$ & $85.2$ & $2.8$   \\
     CMT*~\cite{SGMN13} & $8.7$ & $28.0$ & $13.3$ & $4.7$ & $60.1$ & $8.7$ & $8.4$ & $86.9$ & $15.3$ & $8.7$ & $89.0$ & $15.9$ & $\mathbf{10.9}$ & $74.2$ & $\mathbf{19.0}$   \\   
     SSE~\cite{ZV15} & $2.1$ & $36.4$ & $4.0$ & $8.5$ & $46.9$ & $14.4$ & $7.0$ & $80.5$ & $12.9$ & $8.1$ & $82.5$ & $14.8$ & $0.3$ & $78.4$ & $0.6$   \\
     LATEM~\cite{XASNHS16} & $14.7$ & $28.8$ & $19.5$ & $15.2$ & $57.3$ & $24.0$ & $7.3$ & $71.7$ & $13.3$ & $11.5$ & $77.3$ & $20.0$ & $1.3$ & $71.4$ & $2.6$   \\
     ALE~\cite{APHS15}  & $\mathbf{21.8}$ & $33.1$ & $\mathbf{26.3}$ & $23.7$ & $62.8$ & $\mathbf{34.4}$ & $\mathbf{16.8}$ & $76.1$ & $\mathbf{27.5}$ & $14.0$ & $81.8$ & $23.9$ & $4.6$ & $73.7$ & $8.7$ \\
     DEVISE~\cite{FCSBDRM13}  & $16.9$ & $27.4$ & $20.9$ & $\mathbf{23.8}$ & $53.0$ & $32.8$ & $13.4$ & $68.7$ & $22.4$ & $\mathbf{17.1}$ & $74.7$ &  $\mathbf{27.8}$ & $3.5$ & $78.4$ & $6.7$   \\
     SJE~\cite{ARWLS15} & $14.4$ & $29.7$ & $19.4$ & $23.5$ & $59.2$ & $33.6$ & $11.3$ & $74.6$ & $19.6$ & $8.0$ & $73.9$ & $14.4$ & $1.3$ & $71.4$ & $2.6$   \\
     ESZSL~\cite{RT15} & $11.0$ & $27.9$ & $15.8$ & $14.7$ & $56.5$ & $23.3$ & $6.6$ & $75.6$ & $12.1$ & $5.9$ & $77.8$ & $11.0$ & $2.4$ & $70.1$ & $4.6$   \\
     SYNC~\cite{CCGS16} & $7.9$ & $\mathbf{43.3}$ & $13.4$ & $11.5$  & $\mathbf{70.9}$ & $19.8$ & $9.0$ & $88.9$ & $16.3$ & $9.7$ & $89.7$ & $17.5$ & $7.4$ & $66.3$ & $13.3$ \\
     SAE~\cite{kodirov2017semantic} & $8.8$ & $18.0$ & $11.8$ & $7.8$  & $54.0$ & $13.6$ & $1.8$ & $77.1$ & $3.5$ & $1.1$ & $82.2$ & $2.2$ & $0.4$ & $80.9$ & $0.9$ \\
     GFZSL~\cite{VR17} & $0.0$ & $39.6$ & $0.0$ & $0.0$  & $45.7$ & $0.0$ & $1.8$ & $80.3$ & $3.5$ & $2.5$ & $80.1$ & $4.8$ & $0.0$ & $83.3$ & $0.0$ \\
     \hline
   \end{tabular} 
   \vspace{-3mm}
\caption{Generalized Zero-Shot Learning on Proposed Split Version 2.0 (PS) measuring ts = Top-1 accuracy on $\mathcal{Y}^{ts}$, tr=Top-1 accuracy on $\mathcal{Y}^{tr}$, H = harmonic mean (CMT*: CMT with novelty detection). We measure top-1 accuracy in \%.}
\label{tab:openset}
\end{table*}
}

\myparagraph{Zero-Shot Learning Results on ImageNet.} 
ImageNet scales the methods to a truly large-scale setting, thus these experiments provide further insights on how to tackle the zero-shot learning problem from the practical point of view. Here, we evaluate $10$ methods, i.e. \cite{XASNHS16,ARWLS15,RT15,CCGS16,SGMN13,NMBSSFCD14,FCSBDRM13,APHS15,kodirov2017semantic,VR17}. We exclude DAP and IAP as attributes are not available for all ImageNet classes as well as SSE~\cite{ZV15} due to  scalability issues of the public implementation of the method. \autoref{tab:ImageNet} shows that the best performing method is SYNC~\cite{CCGS16} which may either indicate that it performs well in large-scale setting or it can learn under uncertainty due to usage of Word2Vec instead of attributes. Another possibility is Word2Vec may be tuned for SYNC as it is provided by the same authors. However, we refrain to make a strong claim as this would requires a full evaluation on class embeddings which is out of the scope of this paper. On the other hand, GFZSL~\cite{VR17} which is the best performing model for attribute datasets perform poorly on ImageNet which may indicate that generative models require a strong class embedding space such as attributes to perform well on ZSL task. Note that due to the computational issues, we were not able to obtain results for GFZSL for 3H, M5K, L5K and All 20K classes.

More detailed observations are as follows. The second highest performing method is ESZSL~\cite{RT15} which is one of the linear embedding models that have an implicit regularization mechanism, which seems to be more effective than early stopping as an explicit regularizer. A general observation from the results of all the methods is that in the most populated classes, the results are higher than the least populated classes which indicates that zero-shot learning on fine-grained ImageNet subsets is a more difficult task. Moreover, we conclude that the nature of the test set, e.g. type of the classes being tested, is more important than the number of classes. Therefore, the selection of the test set is an important aspect of zero-shot learning on large-scale datasets. Furthermore, for all methods we consistently observe a large drop in accuracy between 1K and 5K most populated classes which is expected as 5K contains $\approx 6.6$M images, making the problem much more difficult than 1K ($\approx 1624$ images). It is worth to note that, measuring per-image accuracy in this case would lead to higher results if the labels of the highly populated class samples are predicted correctly. Finally, the largest test set, i.e. All 20K, the results are poor for all methods which indicates the difficulty of this problem where there is a large room for improvement.

Several models in the literature evaluate Top-5 and Top-10 as well as Top-1 accuracy on ImageNet. Top-5 and Top-10 accuracy in this case is reasonable as an image usually contains multiple objects however by construction it is associated with a single label in ImageNet. Hence, we provide a comparison of the same $9$ models according to all these three criteria in~\autoref{fig:imagenet_zsl_top1}. We observe that SYNC~\cite{CCGS16} performs significantly better than other methods when the number of images is higher, e.g. 2H, M500, M1K, whereas the gap reduces when the number of images and the number of classes increase, e.g. 3H, L5K and All. In fact, when for All, all the methods perform similarly and poorly which indicates that there is a large room for improvement in this task. In fact, this observation carries on for all three accuracy measures. For Top-5 (middle) and Top-10 (right) accuracy although the numbers are as expected in general higher, the winning model remains as SYNC, significantly for 2H, M500 and M1K whereas the difference is smaller with 3H, L5H, L1K. On the other hand, all methods perform similarly when all 20K classes are tested.

\begin{figure*}[t]
	\centering
        \includegraphics[width=0.28\linewidth,angle=-90, trim=0 30 0 70,clip]{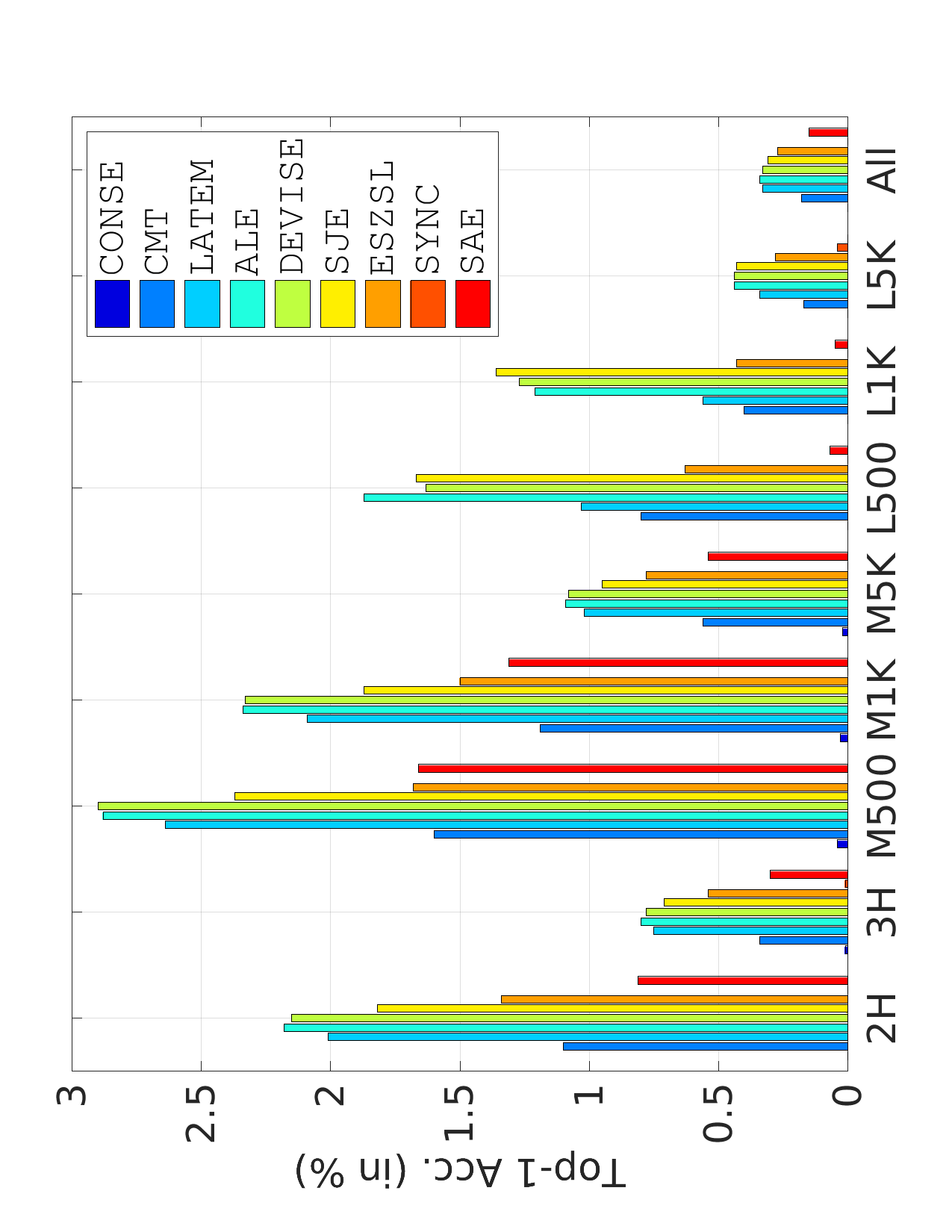}
        \includegraphics[width=0.28\linewidth, angle=-90, trim=0 30 0 70,clip]{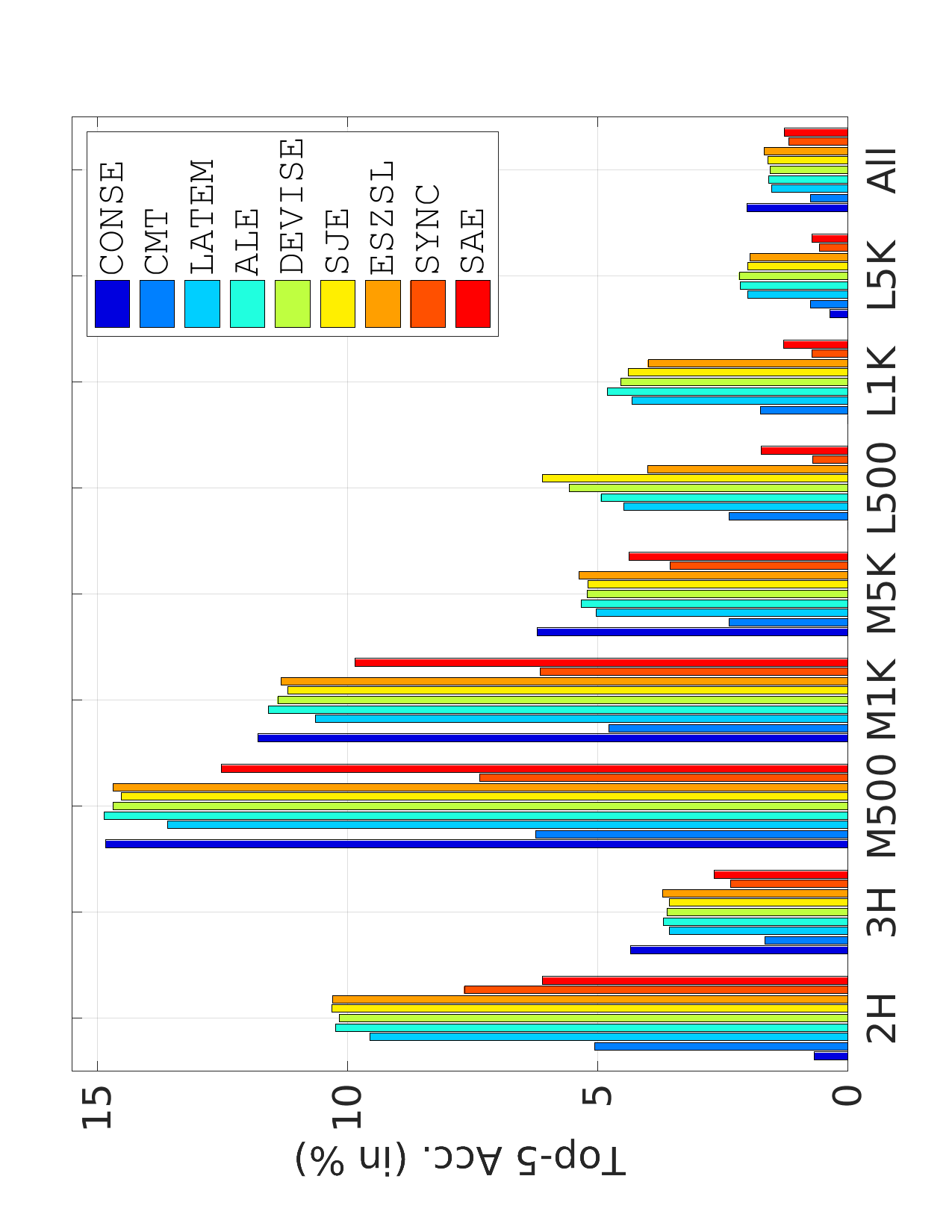}
        \includegraphics[width=0.28\linewidth, angle=-90, trim=0 30 0 70,clip]{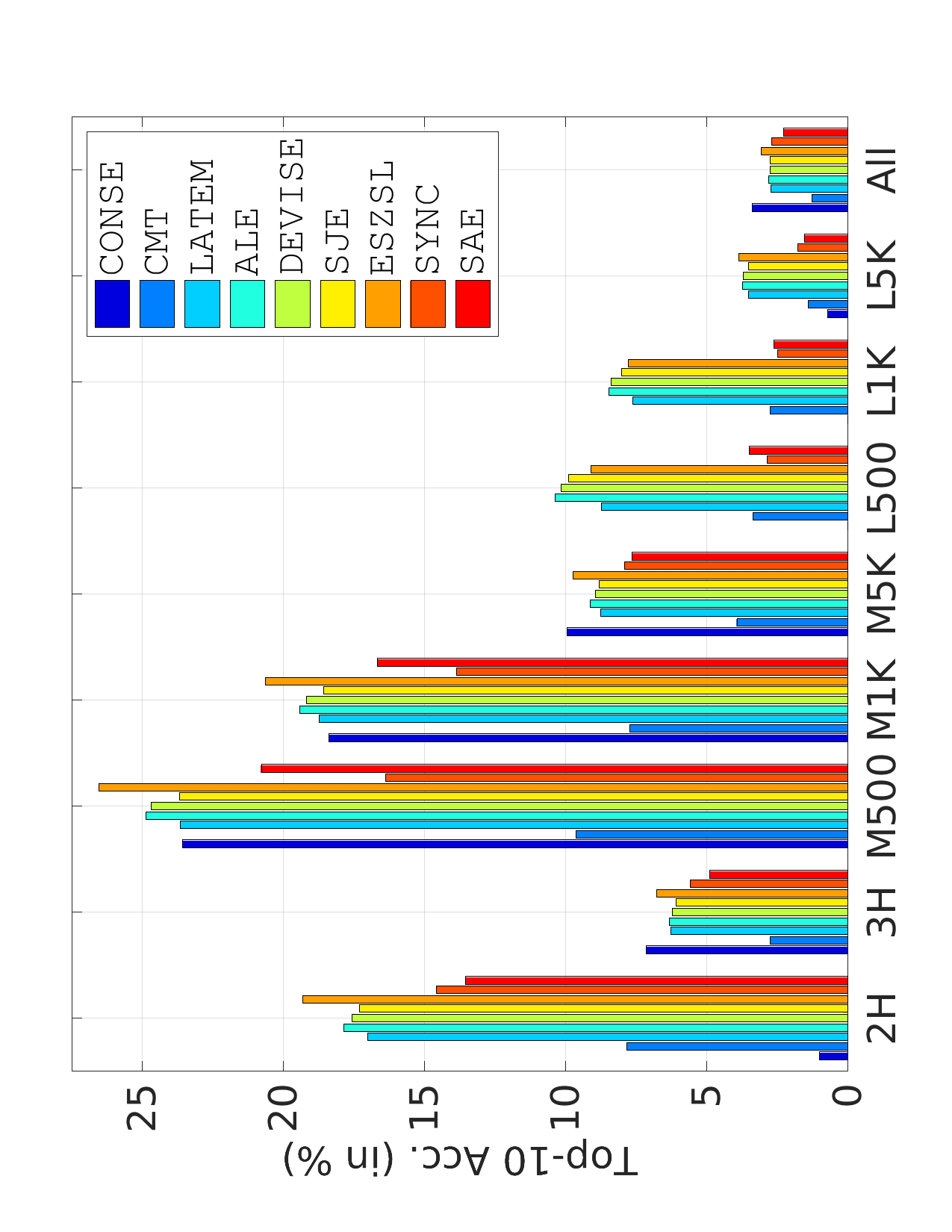}
        \vspace{-5mm}
	\caption{GZSL on Imagenet, measuring Top-1, Top-5 and Top-10 accuracy. 2/3H: classes with 2/3 hops away from ImageNet1K $\mathcal{Y}^{tr}$, M500/M1K/M5K: 500/1K/5K most populated classes, L500/L1K/L5K: 500/1K/5K least populated classes, All: Remaining 20K classes.}
	\label{fig:imagenet_gzsl_top1}
\end{figure*}

\begin{figure}[t]
	\centering 
    \includegraphics[width=0.49\columnwidth, trim=0 60 70 0,clip]{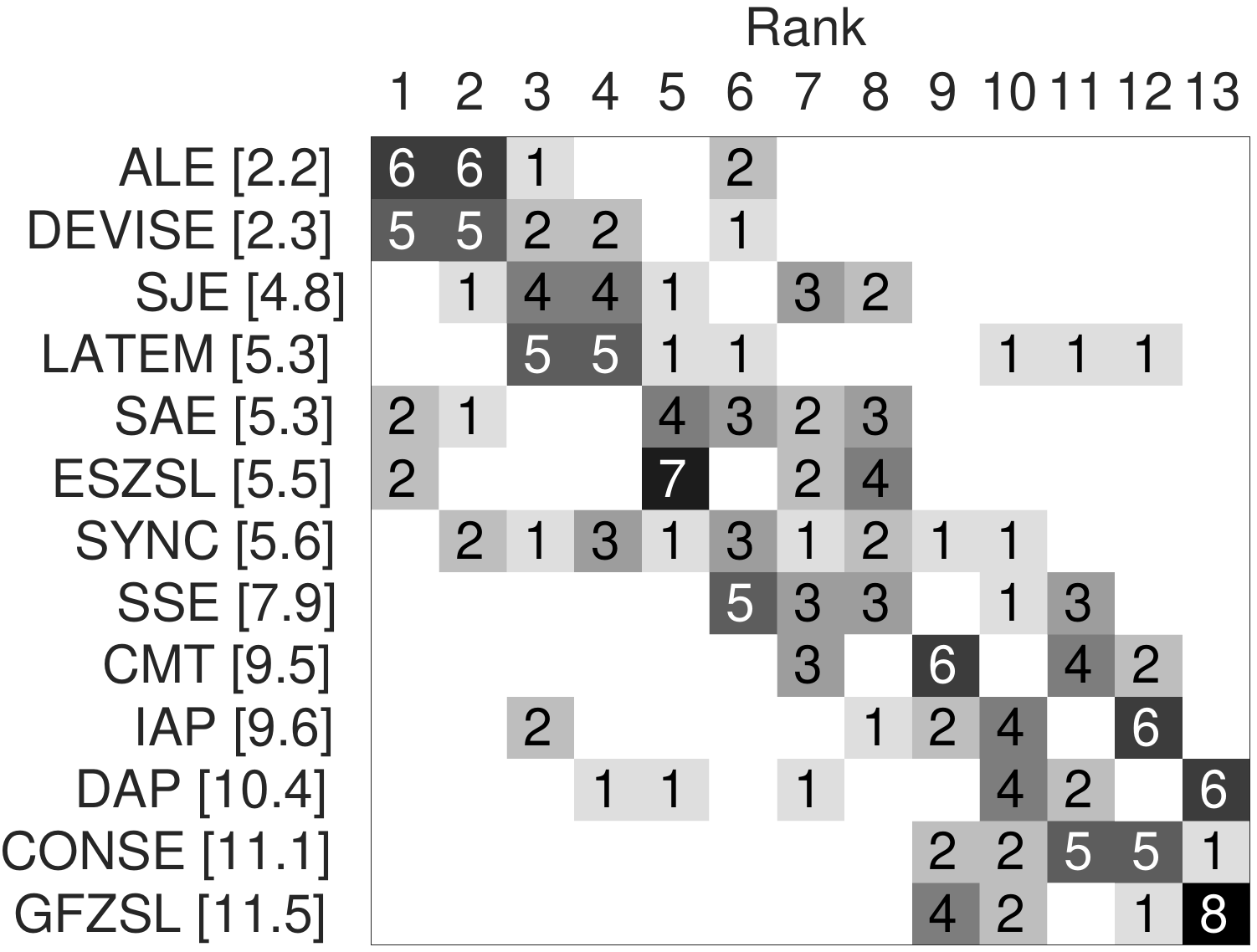}
	\includegraphics[width=0.49\columnwidth, trim=0 60 70 0,clip]{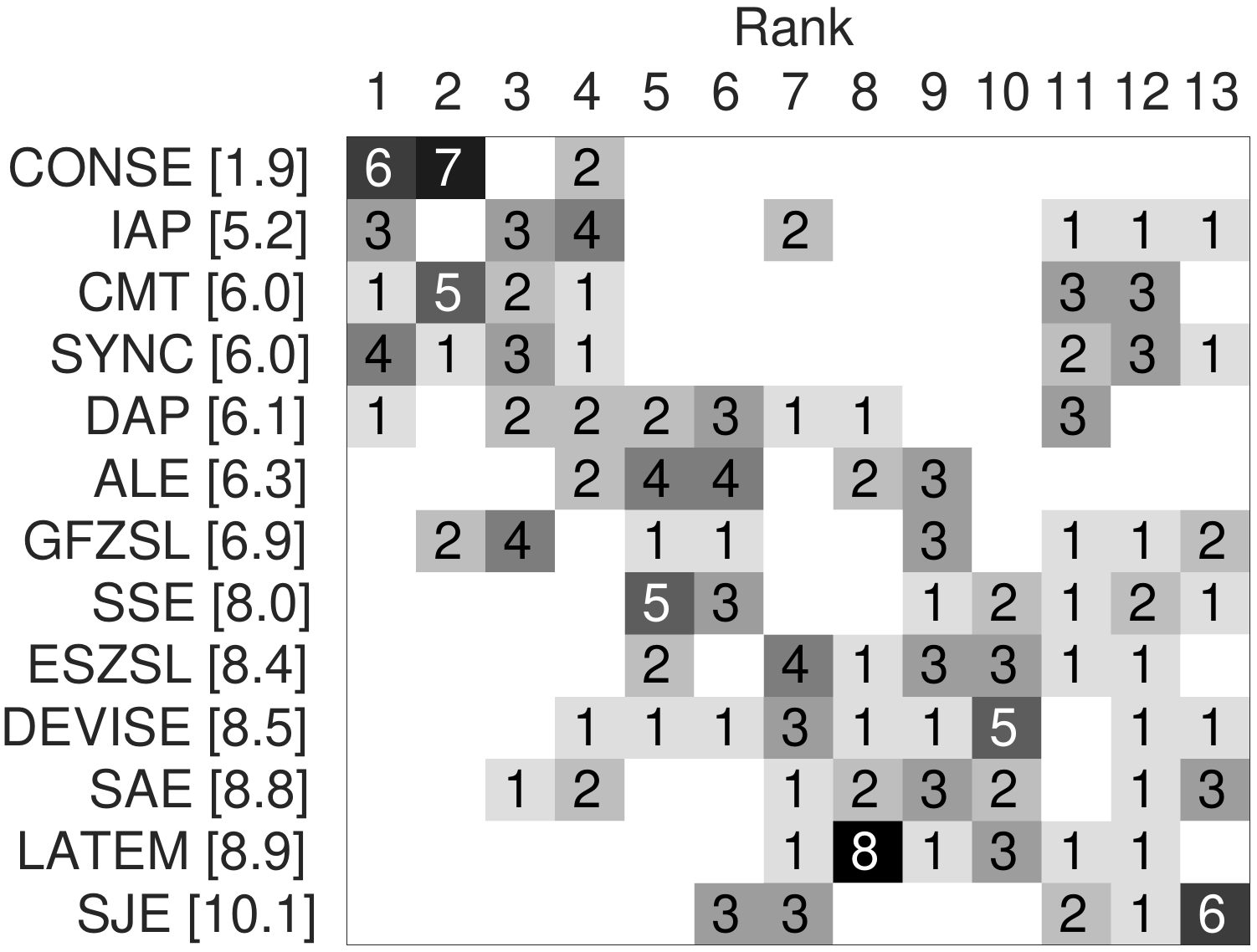} \\ \vspace{2mm}
    \includegraphics[width=0.49\columnwidth, trim=0 60 70 0,clip]{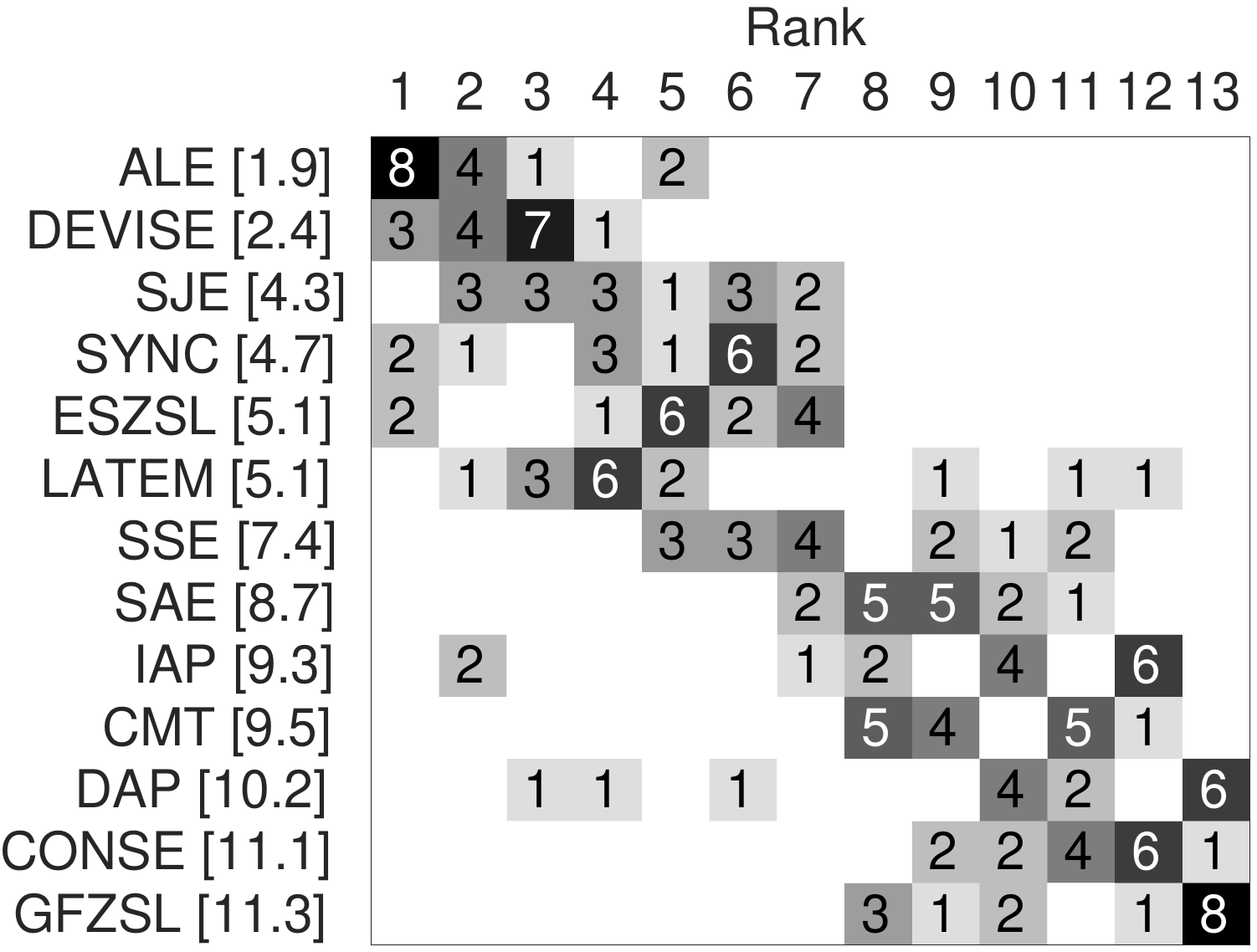}
    \vspace{-2mm}
    \caption{Ranking 13 models on the Proposed Split Version 2.0 in generalized zero-shot learning setting. Top-Left: ranking the Top-1 accuracy (T1) on unseen classes (ts), Top-Right: ranking T1 on seen classes (tr),  Bottom: ranking the Harmonic mean (H).} 
	\label{fig:rankGeneralized}
\end{figure}

\subsection{Generalized Zero-Shot Learning Results}
In real world applications, image classification systems do not have access to whether a novel image belongs to a seen or unseen class in advance. Hence, generalized zero-shot learning is interesting from a practical point of view. Here, we use same models trained on ZSL setting on our proposed splits (PS). We evaluate performance on both $\mathcal{Y}^{tr}$ and $\mathcal{Y}^{ts}$ (using held-out images).

As shown in \autoref{tab:openset}, generalized zero-shot learning results are significantly lower than zero-shot learning results. This is due to the fact that training classes are included in the search space which act as distractors for the images that come from test classes. An interesting observation is that compatibility learning frameworks, e.g. ALE, DEVISE, SJE, perform well on test classes. However, methods that learn independent attribute or object classifiers, e.g. DAP and CONSE, perform well on training classes. Due to this discrepancy, we evaluate the harmonic mean which takes a weighted average of training and test class accuracy as shown in \autoref{eq:H}. The harmonic mean measure ranks ALE as the best performing method on SUN, CUB and AWA1 datasets whereas on our AWA2 dataset DEVISE performs the best and on aPY dataset CMT* performs the best. Note that CMT* has an integrated novelty detection phase for which the method receives another supervision signal determining if the image belongs to a training or a test class. Similar to the ImageNet results, GFZSL~\cite{VR17} performs poorly on GZSL setting.

As for the generalized zero-shot learning setting on ImageNet, we report results measured on unseen classes as no images are reserved from seen classes on \autoref{fig:imagenet_gzsl_top1}. Our first observation is that there is no winner model in all cases, the results diverge for different splits and different accuracy measures. For instance, when the performance is measured with Top-1 accuracy, in general the best performing model seems to be DEVISE, ALE and SJE which are all linear compatibility learning models. On the other hand, for Top-5 accuracy different models take the lead in different splits, e.g. CONSE works the best for 3H and M5K indicating that it performs better when the number of images that come from unseen classes is larger. Whereas SJE and ESZSL works better for 2H, M500, L5H settings. Finally, for Top-10 accuracy, the best performing model overall is ESZSL which is the model that learns a linear compatibility with an explicit regularization scheme. Finally, for Top-1, Top-5 and Top-10 results we observe the same trend for when all the unseen classes are included in the test set, i.e. the models perform similarly however CONSE slightly stands out for Top-5 and Top-10 accuracy plots.

In summary, generalized zero-shot learning setting provides one more level of detail on the performance of zero-shot learning methods. Our take-home message is that the accuracy of training classes is as important as the accuracy of test classes in real world scenarios. Therefore, methods should be designed in a way that they are able to predict labels well both in train and test classes.

\myparagraph{Visualizing Method Ranking.} Similar to the analysis in the previous section that was conducted for zero-shot learning setting, we rank the $13$ methods, i.e. \cite{LNH13,ZV15,XASNHS16,ARWLS15,RT15,CCGS16,SGMN13,NMBSSFCD14,FCSBDRM13,APHS15,kodirov2017semantic,VR17}, based on their results obtained on SUN, CUB, AWA1, AWA2 and aPY. The performance is measured on seen classes, unseen classes and the Harmonic mean of the two. 

The rank matrix of test classes, i.e. \autoref{fig:rankGeneralized} top left, shows that highest ranked methods,i.e. ALE, DEVISE, SJE, although overall the absolute accuracy numbers are lower (\autoref{tab:openset}). Note that in~\autoref{fig:rank_matrix} GFZSL ranked highest which shows that GFZSL is not as strong for GZSL task. The rank matrix of the harmonic mean shows the same trend. However, the rank matrix of training classes, i.e. \autoref{fig:rankGeneralized} top right, shows that models that learn intermediate attribute classifiers perform well for the images that come from training classes. However, these models typically do not lead to a high accuracy for the images that belong to unseen classes as shown in \autoref{tab:openset}. This eventually makes the harmonic mean, i.e. the overall accuracy on both training and test classes, lower. These results clearly suggest that one should not only optimize for test class accuracy but also for training class accuracy while evaluating generalized zero-shot learning. 

Our final observation from~\autoref{fig:rankGeneralized} is that CMT* is better than CMT in all cases which supports the argument that a simple novelty detection scheme helps to improve results. However, it is important to note that the proposed novelty detection mechanism uses more supervision than classic zero-shot learning models. Although the label of test classes is not used, whether the sample comes from a seen or unseen class is an additional supervision.

\subsection{Transductive (Generalized) Zero-Shot Learning}
In contrast to previous zero-shot learning approaches that learn only with data from training classes, transductive approaches use unlabaled images from test classes. In this section, we evaluate three state-of-the-art transductive ZSL approaches, i.e. DSRL~\cite{YG17}, GFZSL-tran~\cite{VR17}, and ALE-tran~\cite{APHS15}. Similar to the previous section, we evaluate those approaches on our proposed splits in both zero-shot learning where test time search space is composed of only unseen classes and generalized zero-shot learning where it contains both seen and unseen classes. The performance is per-class averaged top-1 accuracy.

\begin{figure}[t]
	\centering
        \includegraphics[width=0.4\linewidth,angle=-90, trim=0 10 0 70,clip]{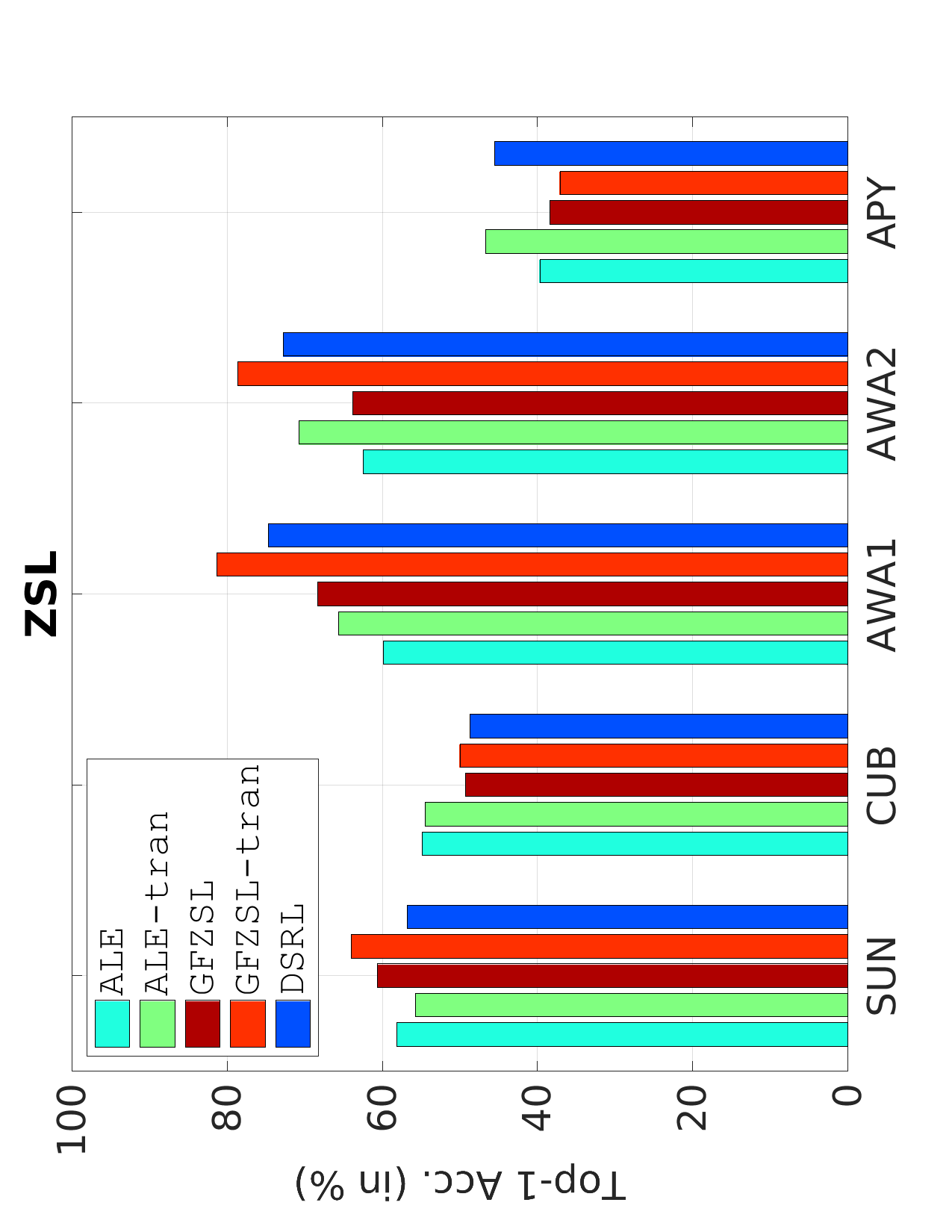}
        \includegraphics[width=0.4\linewidth, angle=-90, trim=0 10 0 70,clip]{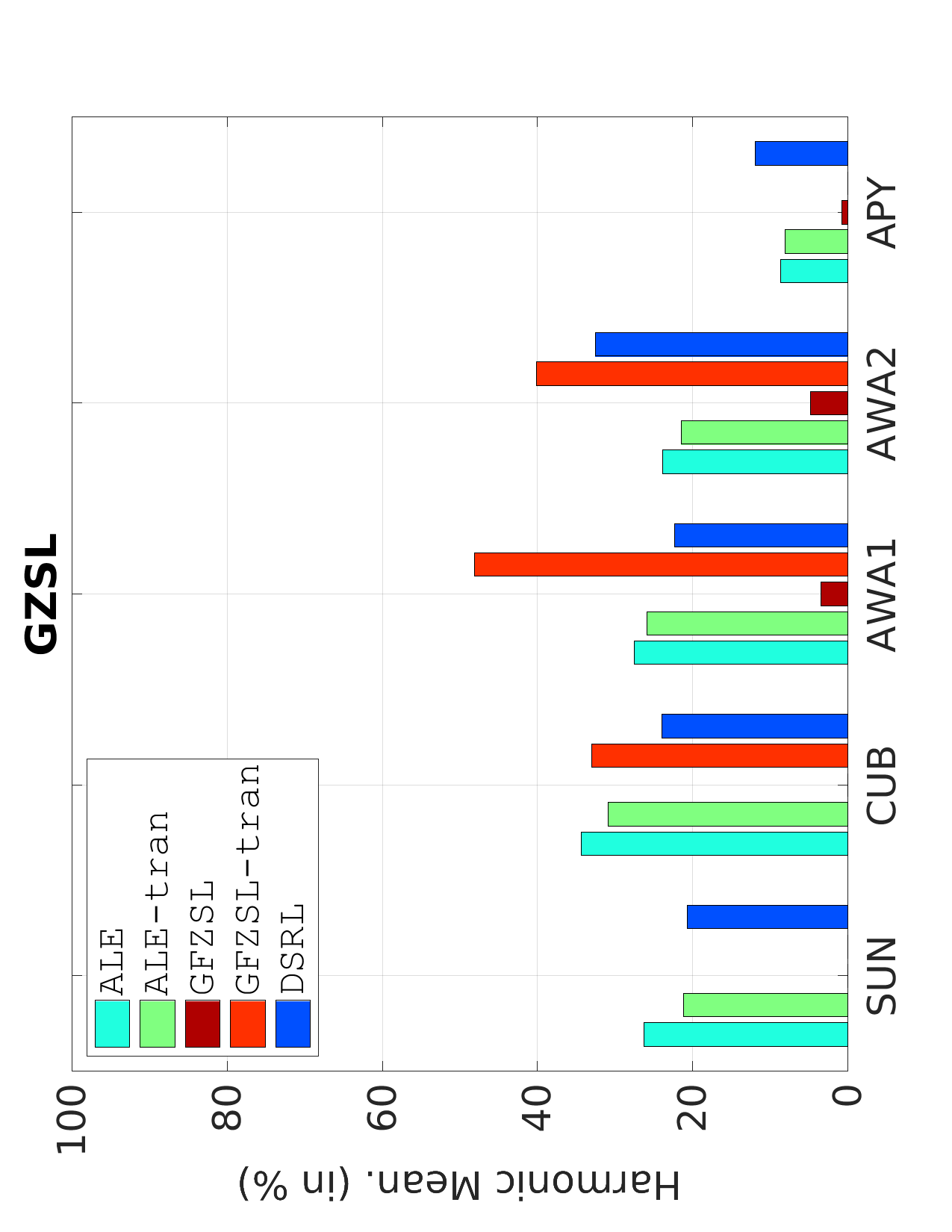}
        \vspace{-5mm}
	\caption{Zero-shot~(left) and generalized zero-shot learning~(right) results in the transductive learning setting on our Proposesd Split.}
	\label{fig:transductive}
\end{figure}

Our transductive learning results are presented in~\autoref{fig:transductive}. We observe that in ZSL setting, transductive learning leads to accuracy improvement, e.g. ALE-tran and GFZSL-tran outperforms ALE and GFZSL respectively in almost all cases. In particular, on AWA2, GFZSL-tran achieves $78.6\%$, significantly improving GFZSL~($63.8\%$). On APY, ALE-tran obtains $45.5\%$ and significantly improves ALE~($37.1\%$) as well. Moreover, GFZSL-tran outperforms ALE-tran and DSRL on SUN, AWA1 and AWA2. However, ALE-tran performs the best on CUB and APY. In GZSL setting we observe a different trend, i.e. transductive learning does not improve results for ALE in any of the datasets. Although, on AWA1 and AWA2 GFZSL results improve significantly for the transductive learning setting, on other datasets GFZSL model performs poorly both in inductive and in transductive settings. 

\section{Conclusion}
\label{sec:conc}
In this work, we evaluated a significant number of state-of-the-art zero-shot learning methods, i.e. \cite{LNH13,ZV15,XASNHS16,ARWLS15,RT15,CCGS16,SGMN13,NMBSSFCD14,FCSBDRM13,APHS15,kodirov2017semantic,VR17,YG17}, on several datasets, i.e. SUN, CUB, AWA1, AWA2, aPY and ImageNet, within a unified evaluation protocol both in zero-shot and generalized zero-shot settings.

Our evaluation showed that generative models and compatibility learning frameworks have an edge over learning independent object or attribute classifiers and also over other hybrid models for the classic zero-shot learning setting. We observed that unlabeled data of unseen classes can further improve the zero-shot learning results, thus it is not fair to compare transductive learning approaches with inductive ones. We discovered that some standard zero-shot dataset splits may treat feature learning disjoint from the training stage as several test classes are included in the ImageNet1K dataset that is used to train the deep neural networks that act as feature extractor. Therefore, we proposed new dataset splits making sure that none of the test classes in none of the datasets belong to ImageNet1K. Moreover, disjoint training and validation class split is a necessary component of parameter tuning in zero-shot learning setting. 

In addition, we introduced a new Animal with Attributes (AWA2) dataset. AWA2 inherits the same 50 classes and attributes annotations from the original Animal with Attributes (AWA1) dataset, but consists of different $37,322$ images with publicly available redistribution license. Our experimental results showed that the 12 methods that we evaluated perform similarly on AWA2 and AWA1. Moreover, our statistical consistency test indicated that AWA1 and AWA2 are compatible with each other.

Finally, including training classes in the search space while evaluating the methods, i.e. generalized zero-shot learning, provides an interesting playground for future research. Although the generalized zero-shot learning accuracy obtained with 13 models compared to their zero-shot learning accuracy is significantly lower, the relative performance comparison of different models remain the same. Having noticed that some models perform well when the test set is composed only of seen classes, while some others perform well when the test set is composed of only of unseen classes, we proposed the Harmonic mean of seen and unseen class accuracy as a unified measure for performance in GZSL setting. The Harmonic mean encourages the models to perform well on both seen and unseen class samples, which is closer to a real world setting. In summary, our work extensively evaluated the good and bad aspects of zero-shot learning while sanitizing the ugly ones.

{
\small
	\bibliographystyle{IEEEtran}
	\bibliography{biblio}

\begin{thebibliography}{10}
\providecommand{\url}[1]{#1}
\csname url@samestyle\endcsname
\providecommand{\newblock}{\relax}
\providecommand{\bibinfo}[2]{#2}
\providecommand{\BIBentrySTDinterwordspacing}{\spaceskip=0pt\relax}
\providecommand{\BIBentryALTinterwordstretchfactor}{4}
\providecommand{\BIBentryALTinterwordspacing}{\spaceskip=\fontdimen2\font plus
\BIBentryALTinterwordstretchfactor\fontdimen3\font minus
  \fontdimen4\font\relax}
\providecommand{\BIBforeignlanguage}[2]{{%
\expandafter\ifx\csname l@#1\endcsname\relax
\typeout{** WARNING: IEEEtran.bst: No hyphenation pattern has been}%
\typeout{** loaded for the language `#1'. Using the pattern for}%
\typeout{** the default language instead.}%
\else
\language=\csname l@#1\endcsname
\fi
#2}}
\providecommand{\BIBdecl}{\relax}
\BIBdecl

\bibitem{LNH13}
C.~Lampert, H.~Nickisch, and S.~Harmeling, ``Attribute-based classification for
  zero-shot visual object categorization,'' in \emph{TPAMI}, 2013.

\bibitem{LEB08}
H.~Larochelle, D.~Erhan, and Y.~Bengio, ``Zero-data learning of new tasks,'' in
  \emph{AAAI}, 2008.

\bibitem{RSS11}
M.~Rohrbach, M.~Stark, and B.Schiele, ``Evaluating knowledge transfer and
  zero-shot learning in a large-scale setting,'' in \emph{CVPR}, 2011.

\bibitem{YA10}
X.~Yu and Y.~Aloimonos, ``Attribute-based transfer learning for object
  categorization with zero or one training example,'' in \emph{ECCV}, 2010.

\bibitem{Xu17}
X.~Xu, Y.~Yang, D.~Zhang, H.~T. Shen, and J.~Song, ``Matrix tri-factorization
  with manifold regularizations for zero-shot learning,'' in \emph{CVPR}, 2017.

\bibitem{Ding17}
Z.~Ding, M.~Shao, and Y.~Fu, ``Low-rank embedded ensemble semantic dictionary
  for zero-shot learning,'' in \emph{CVPR}, 2017.

\bibitem{FCSBDRM13}
A.~Frome, G.~S. Corrado, J.~Shlens, S.~Bengio, J.~Dean, M.~A. Ranzato, and
  T.~Mikolov, ``Devise: A deep visual-semantic embedding model,'' in
  \emph{NIPS}, 2013, pp. 2121--2129.

\bibitem{APHS13}
Z.~Akata, F.~Perronnin, Z.~Harchaoui, and C.~Schmid, ``{Label embedding for
  attribute-based classification},'' in \emph{CVPR}, 2013.

\bibitem{ARWLS15}
Z.~Akata, S.~Reed, D.~Walter, H.~Lee, and B.~Schiele, ``Evaluation of output
  embeddings for fine-grained image classification,'' in \emph{CVPR}, 2015.

\bibitem{RT15}
B.~Romera-Paredes and P.~H. Torr, ``An embarrassingly simple approach to
  zero-shot learning,'' \emph{ICML}, 2015.

\bibitem{XASNHS16}
Y.~Xian, Z.~Akata, G.~Sharma, Q.~Nguyen, M.~Hein, and B.~Schiele, ``Latent
  embeddings for zero-shot classification,'' in \emph{CVPR}, 2016.

\bibitem{SGMN13}
R.~Socher, M.~Ganjoo, C.~D. Manning, and A.~Ng, ``Zero-shot learning through
  cross-modal transfer,'' in \emph{NIPS}, 2013.

\bibitem{ZV15}
Z.~Zhang and V.~Saligrama, ``Zero-shot learning via semantic similarity
  embedding,'' in \emph{ICCV}, 2015.

\bibitem{CCGS16}
S.~Changpinyo, W.-L. Chao, B.~Gong, and F.~Sha, ``Synthesized classifiers for
  zero-shot learning,'' in \emph{CVPR}, 2016.

\bibitem{NMBSSFCD14}
M.~Norouzi, T.~Mikolov, S.~Bengio, Y.~Singer, J.~Shlens, A.~Frome, G.~Corrado,
  and J.~Dean, ``Zero-shot learning by convex combination of semantic
  embeddings,'' in \emph{ICLR}, 2014.

\bibitem{PH12}
G.~Patterson and J.~Hays, ``Sun attribute database: Discovering, annotating,
  and recognizing scene attributes,'' in \emph{CVPR}, 2012.

\bibitem{CaltechUCSDBirdsDataset}
P.~Welinder, S.~Branson, T.~Mita, C.~Wah, F.~Schroff, S.~Belongie, and
  P.~Perona, ``{Caltech-UCSD Birds 200},'' Caltech, Tech. Rep. CNS-TR-2010-001,
  2010.

\bibitem{FEHF09}
A.~Farhadi, I.~Endres, D.~Hoiem, and D.~Forsyth, ``Describing objects by their
  attributes,'' \emph{CVPR}, 2009.

\bibitem{ImageNet}
J.~Deng, W.~Dong, R.~Socher, L.-J. Li, K.~Li, and L.~Fei-Fei, ``{ImageNet: A
  Large-Scale Hierarchical Image Database},'' in \emph{CVPR}, 2009.

\bibitem{lowe2004distinctive}
D.~G. Lowe, ``Distinctive image features from scale-invariant keypoints,''
  \emph{IJCV}, vol.~60, no.~2, pp. 91--110, 2004.

\bibitem{donahue2014decaf}
J.~Donahue, Y.~Jia, O.~Vinyals, J.~Hoffman, N.~Zhang, E.~Tzeng, and T.~Darrell,
  ``Decaf: A deep convolutional activation feature for generic visual
  recognition.'' in \emph{ICML}, 2014.

\bibitem{simonyan2014very}
K.~Simonyan and A.~Zisserman, ``Very deep convolutional networks for
  large-scale image recognition,'' \emph{arXiv preprint arXiv:1409.1556}, 2014.

\bibitem{HZRS15}
K.~He, X.~Zhang, S.~Ren, and J.~Sun, ``Deep residual learning for image
  recognition,'' in \emph{CVPR}, 2016.

\bibitem{HTS16}
Z.~Al-Halah, M.~Tapaswi, and R.~Stiefelhagen, ``Recovering the missing link:
  Predicting class-attribute associations for unsupervised zero-shot
  learning,'' in \emph{CVPR}, 2016.

\bibitem{jayaraman2014zero}
D.~Jayaraman and K.~Grauman, ``Zero-shot recognition with unreliable
  attributes,'' in \emph{NIPS}, 2014.

\bibitem{KKTH12}
P.~Kankuekul, A.~Kawewong, S.~Tangruamsub, and O.~Hasegawa, ``Online
  incremental attribute-based zero-shot learning,'' in \emph{CVPR}, 2012.

\bibitem{MSCCD13}
T.~Mikolov, I.~Sutskever, K.~Chen, G.~S. Corrado, and J.~Dean, ``Distributed
  representations of words and phrases and their compositionality,'' in
  \emph{NIPS}, 2013.

\bibitem{FHXG15}
Y.~Fu, T.~M. Hospedales, T.~Xiang, and S.~Gong, ``Transductive multi-view
  zero-shot learning,'' \emph{TPAMI}, 2015.

\bibitem{palatucci2009zero}
M.~Palatucci, D.~Pomerleau, G.~E. Hinton, and T.~M. Mitchell, ``Zero-shot
  learning with semantic output codes,'' in \emph{NIPS}, 2009.

\bibitem{APHS15}
Z.~Akata, F.~Perronnin, Z.~Harchaoui, and C.~Schmid, ``Label-embedding for
  image classification,'' \emph{TPAMI}, 2016.

\bibitem{QLSH16}
R.~Qiao, L.~Liu, C.~Shen, and A.~van~den Hengel, ``Less is more: Zero-shot
  learning from online textual documents with noise suppression,'' in
  \emph{CVPR}, 2016.

\bibitem{BHJ16}
M.~Bucher, S.~Herbin, and F.~Jurie, ``Improving semantic embedding consistency
  by metric learning for zero-shot classiffication,'' in \emph{ECCV}, 2016, pp.
  730--746.

\bibitem{kodirov2017semantic}
E.~Kodirov, T.~Xiang, and S.~Gong, ``Semantic autoencoder for zero-shot
  learning,'' in \emph{CVPR}, 2017.

\bibitem{lei2015predicting}
J.~Lei~Ba, K.~Swersky, S.~Fidler \emph{et~al.}, ``Predicting deep zero-shot
  convolutional neural networks using textual descriptions,'' in \emph{ICCV},
  2015.

\bibitem{zhang2016learning}
L.~Zhang, T.~Xiang, and S.~Gong, ``Learning a deep embedding model for
  zero-shot learning,'' in \emph{CVPR}, 2017.

\bibitem{changpinyo2016predicting}
S.~Changpinyo, W.-L. Chao, and F.~Sha, ``Predicting visual exemplars of unseen
  classes for zero-shot learning,'' \emph{ICCV}, pp. 3496--3505, 2017.

\bibitem{ZV16}
Z.~Zhang and V.~Saligrama, ``Zero-shot learning via joint semantic similarity
  embedding,'' in \emph{CVPR}, 2016.

\bibitem{FXKG15}
Z.~Fu, T.~Xiang, E.~Kodirov, and S.~Gong, ``Zero-shot object recognition by
  semantic manifold distance,'' in \emph{CVPR}, June 2015.

\bibitem{AMFS16}
Z.~Akata, M.~Malinowski, M.~Fritz, and B.~Schiele, ``Multi-cue zero-shot
  learning with strong supervision,'' in \emph{CVPR}, 2016.

\bibitem{long2017zero}
Y.~Long, L.~Liu, L.~Shao, F.~Shen, G.~Ding, and J.~Han, ``From zero-shot
  learning to conventional supervised classification: Unseen visual data
  synthesis,'' in \emph{CVPR}, 2017.

\bibitem{VR17}
V.~K. Verm and P.~Rai, ``A simple exponential family framework for zero-shot
  learning,'' in \emph{ECML}, 2017, pp. 792--808.

\bibitem{LW17}
Y.~Li and D.~Wang, ``Zero-shot learning with generative latent prototype
  model,'' \emph{arXiv preprint arXiv:1705.09474}, 2017.

\bibitem{MH16}
T.~Mukherjee and T.~Hospedales, ``Gaussian visual-linguistic embedding for
  zero-shot recognition,'' in \emph{EMNLP}, 2016.

\bibitem{MES13}
M.~Rohrbach, S.~Ebert, and B.~Schiele, ``Transfer learning in a transductive
  setting,'' in \emph{NIPS}, 2013.

\bibitem{KXFG15}
E.~Kodirov, T.~Xiang, Z.~Fu, and S.~Gong, ``Unsupervised domain adaptation for
  zero-shot learning,'' in \emph{ICCV}, 2015.

\bibitem{LGS15}
X.~Li, Y.~Guo, and D.~Schuurmans, ``Semi-supervised zero-shot classification
  with label representation learning,'' in \emph{ICCV}, 2015.

\bibitem{li2015max}
X.~Li and Y.~Guo, ``Max-margin zero-shot learning for multi-class
  classification.'' in \emph{AISTATS}, 2015.

\bibitem{FS16}
Y.~Fu and L.~Sigal, ``Semi-supervised vocabulary-informed learning,'' in
  \emph{CVPR}, 2016.

\bibitem{ZV16b}
Z.~Zhang and V.~Saligrama, ``Zero-shot recognition via structured prediction,''
  in \emph{CVPR}, 2016.

\bibitem{MGS14}
T.~Mensink, E.~Gavves, and C.~G. Snoek, ``Costa: Co-occurrence statistics for
  zero-shot classification,'' in \emph{CVPR}, 2014.

\bibitem{rohrbach2010helps}
M.~Rohrbach, M.~Stark, G.~Szarvas, I.~Gurevych, and B.~Schiele, ``What helps
  where--and why? semantic relatedness for knowledge transfer,'' in
  \emph{CVPR}, 2010.

\bibitem{RALS16}
S.~Reed, Z.~Akata, H.~Lee, and B.~Schiele, ``Learning deep representations of
  fine-grained visual descriptions,'' in \emph{CVPR}, 2016.

\bibitem{ESE13}
M.~Elhoseiny, B.~Saleh, and A.~Elgammal, ``Write a classifier: Zero-shot
  learning using purely textual descriptions,'' in \emph{ICCV}, 2013.

\bibitem{antol2014zero}
S.~Antol, C.~L. Zitnick, and D.~Parikh, ``Zero-shot learning via visual
  abstraction,'' in \emph{European Conference on Computer Vision}.\hskip 1em
  plus 0.5em minus 0.4em\relax Springer, 2014, pp. 401--416.

\bibitem{mensink2012metric}
T.~Mensink, J.~Verbeek, F.~Perronnin, and G.~Csurka, ``Metric learning for
  large scale image classification: Generalizing to new classes at near-zero
  cost,'' \emph{Computer Vision--ECCV 2012}, pp. 488--501, 2012.

\bibitem{PSM14}
J.~Pennington, R.~Socher, and C.~D. Manning, ``Glove: Global vectors for word
  representation,'' in \emph{EMNLP}, 2014, pp. 1532--1543.

\bibitem{WordNet}
\BIBentryALTinterwordspacing
G.~A. Miller, ``Wordnet: a lexical database for english,'' \emph{CACM},
  vol.~38, pp. 39--41, 1995. [Online]. Available:
  \url{http://doi.acm.org/10.1145/219717.219748}
\BIBentrySTDinterwordspacing

\bibitem{KASB17}
N.~Karessli, Z.~Akata, B.~Schiele, and A.~Bulling, ``Gaze embeddings for
  zero-shot image classification,'' in \emph{IEEE Computer Vision and Pattern
  Recognition (CVPR)}, 2017.

\bibitem{SRSB13}
W.~J. Scheirer, A.~Rocha, A.~Sapkota, and T.~E. Boult, ``Towards open set
  recognition,'' \emph{TPAMI}, vol.~36, 2013.

\bibitem{JSB14}
L.~Jain, W.~Scheirer, and T.~Boult, ``Multi-class open set recognition using
  probability of inclusion,'' in \emph{ECCV}, 2014.

\bibitem{ZSYXLC16}
H.~Zhang, X.~Shang, W.~Yang, H.~Xu, H.~Luan, and T.-S. Chua, ``Online
  collaborative learning for open-vocabulary visual classifiers,'' in
  \emph{CVPR}, 2016.

\bibitem{BB16}
A.~Bendale and T.~E. Boult, ``Towards open set deep networks,'' in \emph{CVPR},
  2016.

\bibitem{CCGS16b}
W.-L. Chao, S.~Changpinyo, B.~Gong, and F.~Sha, ``An empirical study and
  analysis of generalized zero-shot learning for object recognition in the
  wild,'' in \emph{ECCV}, 2016.

\bibitem{J02}
T.~Joachims, ``Optimizing search engines using clickthrough data,'' in
  \emph{KDD}.\hskip 1em plus 0.5em minus 0.4em\relax ACM, 2002.

\bibitem{UBG09}
N.~Usunier, D.~Buffoni, and P.~Gallinari, ``Ranking with ordered weighted
  pairwise classification,'' in \emph{ICML}, 2009.

\bibitem{WBU11}
J.~Weston, S.~Bengio, and N.~Usunier, ``Wsabie: Scaling up to large vocabulary
  image annotation,'' in \emph{IJCAI}, 2011.

\bibitem{TJH05}
I.~Tsochantaridis, T.~Joachims, T.~Hofmann, and Y.~Altun, ``Large margin
  methods for structured and interdependent output variables,'' \emph{JMLR},
  2005.

\bibitem{bartels1972solution}
R.~H. Bartels and G.~Stewart, ``Solution of the matrix equation ax+ xb= c
  [f4],'' \emph{Commun. ACM}, vol.~15, no.~9, pp. 820--826, 1972.

\bibitem{chapelle2009semi}
O.~Chapelle, B.~Scholkopf, and A.~Zien, ``Semi-supervised learning,''
  \emph{IEEE Transactions on Neural Networks}, vol.~20, no.~3, pp. 542--542,
  2009.

\bibitem{zhou2004learning}
D.~Zhou, O.~Bousquet, T.~N. Lal, J.~Weston, and B.~Sch{\"o}lkopf, ``Learning
  with local and global consistency,'' in \emph{NIPS}, 2004.

\bibitem{YG17}
M.~Ye and Y.~Guo, ``Zero-shot classification with discriminative semantic
  representation learning,'' in \emph{CVPR}, 2017.

\bibitem{fujiwara2014efficient}
Y.~Fujiwara and G.~Irie, ``Efficient label propagation,'' in \emph{ICML}, 2014,
  pp. 784--792.

\bibitem{SLJSRAEVR15}
C.~Szegedy, W.~Liu, Y.~Jia, P.~Sermanet, S.~Reed, D.~Anguelov, D.~Erhan,
  V.~Vanhoucke, and A.~Rabinovich, ``Going deeper with convolutions,'' in
  \emph{CVPR}, 2015.

\bibitem{zhou2014learning}
B.~Zhou, A.~Lapedriza, J.~Xiao, A.~Torralba, and A.~Oliva, ``Learning deep
  features for scene recognition using places database,'' in \emph{NIPS}, 2014.

\bibitem{szegedy2014going}
C.~Szegedy, W.~Liu, Y.~Jia, P.~Sermanet, S.~Reed, D.~Anguelov, D.~Erhan,
  V.~Vanhoucke, and A.~Rabinovich, ``Going deeper with convolutions,''
  \emph{CVPR}, 2015.

\bibitem{GH08}
S.~Garcia and F.~Herrera, ``An extension on``statistical comparisons of
  classifiers over multiple data sets''for all pairwise comparisons,''
  \emph{JLMR}, vol.~9, pp. 2677--2694, 2008.

\end{thebibliography}
}

\begin{IEEEbiography}[{\includegraphics[width=1in,height=1.2in,clip,keepaspectratio]{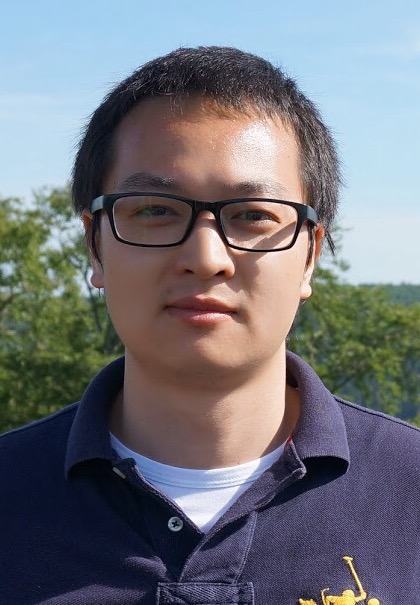}}]{Yongqin Xian}
received his B.Sc. degree in software engineering from Beijing Institute of Technology, China, in 2013 and his M.Sc. degree with honors in computer science from Saarland University, Germany in 2016. He is currently a PhD student in the Computer Vision and Multi-modal Computing group at the Max-Planck Institute for Informatics in Saarbr\"ucken, Germany. His research interests include machine learning and computer vision.
\end{IEEEbiography} 
\vspace{-10mm}
\begin{IEEEbiography}[{\includegraphics[width=1in,height=1.2in,clip,keepaspectratio]{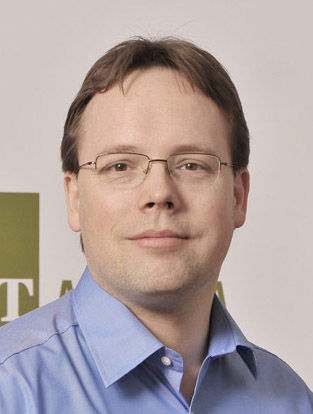}}]{Christoph Lampert}
received the PhD degree in mathematics from the University of Bonn in 2003. In 2010 he joined the Institute of Science and Technology Austria (IST Austria) first as an Assistant Professor and since 2015 as a Professor. His research on computer vision and machine learning has won several international and national awards, including the best paper prize at CVPR 2008. In 2012 he was awarded an ERC Starting Grant by the European Research Council. He is an Editor of the International Journal of Computer Vision (IJCV), Action Editor of the Journal for Machine Learning Research (JMLR), and Associate Editor in Chief of the IEEE Transactions on Pattern Analysis and Machine Intelligence (TPAMI). 
\end{IEEEbiography} 
\vspace{-10mm}
\begin{IEEEbiography}[{\includegraphics[width=1in,height=1.2in,clip,keepaspectratio]{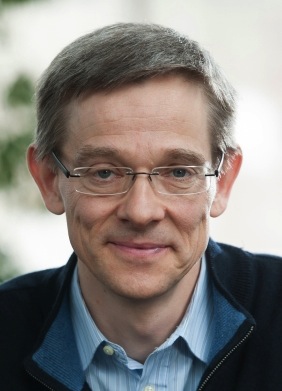}}]{Bernt Schiele} received masters degree in computer science from the University of Karlsruhe and INP Grenoble in 1994 and the PhD degree from INP Grenoble in computer vision in 1997. He was a postdoctoral associate and visiting assistant professor with MIT between 1997 and 2000. From 1999 until 2004, he was an assistant professor with ETH Zurich and, from 2004 to 2010, he was a full professor of computer science with TU Darmstadt. In 2010, he was appointed a scientific member of the Max Planck Society and director at the Max Planck Institute for Informatics. Since 2010, he has also been a professor at Saarland University. His main interests are computer vision, perceptual computing, statistical learning methods, wearable computers, and integration of multimodal sensor data. He is particularly interested in developing methods which work under real world conditions.
\end{IEEEbiography} 
\vspace{-10mm}
\begin{IEEEbiography}[{\includegraphics[width=1in,height=1.2in,clip,keepaspectratio]{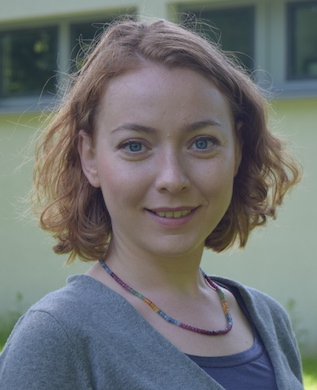}}]{Zeynep Akata}
holds a MSc degree from RWTH Aachen in 2010 and a PhD degree from INRIA of Universit\'e de Grenoble in 2014. She was a post-doctoral researcher in Max Planck Institute for Informatics between 2014-2017 and a visiting researcher at UC Berkeley in 2016-2017. In 2014, she received Lise-Meitner Award for Excellent Women in Computer Science from Max Planck Society. She is currently an assistant professor at the University of Amsterdam, Scientific Manager of the Delta Lab and a Senior Researcher at Max Planck Institute for Informatics. Her research interests include machine learning combined with vision and language for the task of explainable artificial intelligence (XAI).
\end{IEEEbiography} 
\end{document}